\documentclass[]{style}
\usepackage[most]{tcolorbox}        
\usepackage{amsmath,amssymb}
\usepackage{lipsum}                 
\usepackage{xcolor}                 
\usepackage{fancyhdr}               
\usepackage{comment}
\usepackage{float}
\usepackage{booktabs}
\usepackage[table]{xcolor}
\usepackage{url}
\usepackage{graphicx}   
\usepackage{subcaption} 
\usepackage{multirow}
\usepackage{bm}
\definecolor{lightpurple}{RGB}{153, 102, 204}  
\definecolor{lilac}{RGB}{182, 133, 210}        
\usepackage{hyperref}
\hypersetup{
    colorlinks=true,      
    linkcolor=purple,        
    citecolor=lilac,       
    urlcolor=magenta      
}

\definecolor{abstractpurple}{HTML}{9C27B0} 

\definecolor{checkgreen}{RGB}{0, 150, 0}    
\definecolor{crossred}{RGB}{200, 0, 0}      
\newcommand{\cmark}{\textcolor{checkgreen}{\checkmark}}
\newcommand{\xmark}{\textcolor{crossred}{\times}}

\renewcommand{\maketitle}{\mymaketitle}

\begin{document}


\title{Wuying-Browser-Agent: Real-World Centric Fundamental Long-Horizon Browser Agents}
\author{AI Model Application \& Engineering Team,\\
End-User Intelligent Computing BU, Alibaba Cloud }
\affiliation[]{See \hyperref[sec:contrib_ack]{\textcolor{thupurple}{Contributions and Acknowledgments}} section for a full author list.}

\abstract{
Browser agents perform well on short, clean demonstrations, but real deployment is fundamentally different: agents must sustain dozens of decisions on live websites while recovering from mistakes and navigating complex UIs. We argue that closing this gap requires alignment at every level of the pipeline, including execution, supervision, optimization, and evaluation, rather than scale alone. We present Wuying-Browser-Agent, a unified framework that addresses each of these levels. A structured browser harness provides stable execution primitives and decision-oriented context management. Reflection and UI-specialized Curriculum SFT (RUIC-SFT) explicitly trains on recovery trajectories and complex-UI interactions. Divergence-Aware Online GRPO (DAO-GRPO) improves long-horizon credit assignment through potential-based reward shaping and divergence-aware step weighting. Finally, we introduce BrowserBench, a bilingual real-web benchmark of 350 tasks averaging 37.9 steps, because most existing benchmarks are too short to expose long-horizon failure modes. Wuying-Browser-Agent-27B achieves 80.6\% on WebVoyager, 66.7\% on Online-Mind2Web, and 65.1\% on BrowserBench, establishing a new open-source state of the art on browser-use benchmarks. The same pipeline also transfers beyond browser use, demonstrating strong general agentic ability and reaching an average score of 73.8 on Tau2-Bench, Claw-Eval, and BFCL-v4.
}




\maketitle


\begin{figure*}[h]
    \centering
    \includegraphics[width=0.99\textwidth,height=0.38\textwidth]{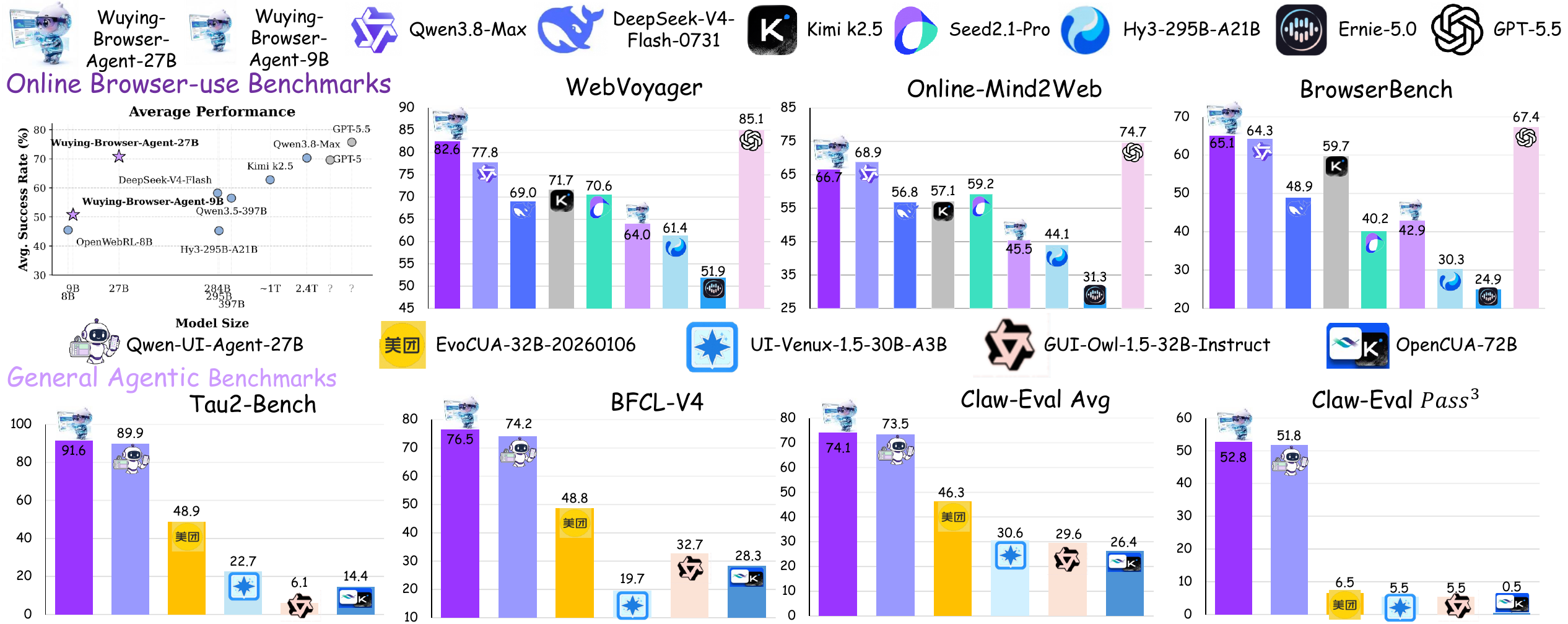}
    \caption{
Performance comparison on online browser-use benchmarks and general agentic benchmarks.
}
    \label{fig:compare}
\end{figure*}

\section{Introduction}\label{sec:introduction}



\begin{figure*}[t]
    \centering
    \includegraphics[width=0.99\textwidth]{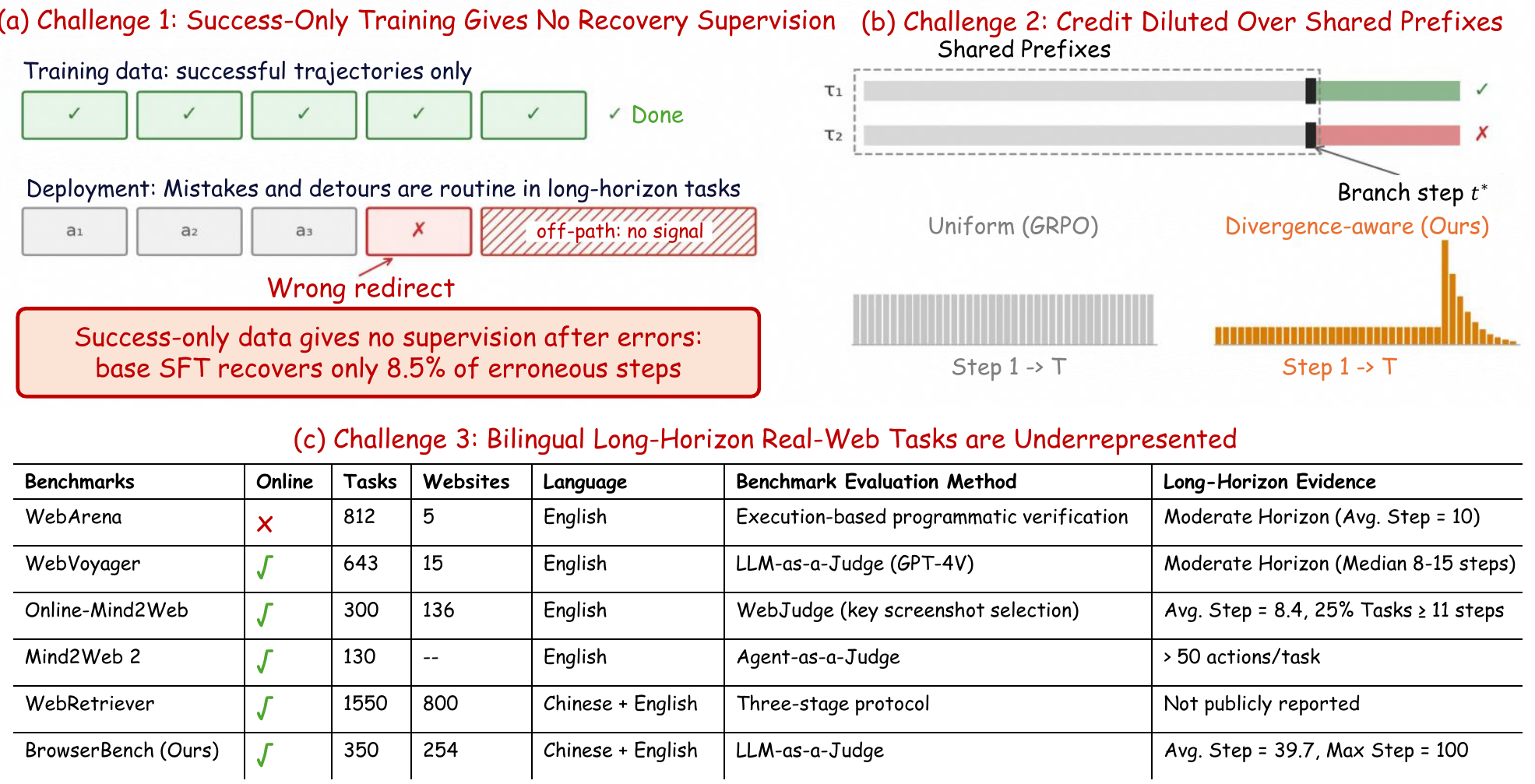}
    \caption{
Three structural challenges for browser agents in the long-horizon regime, and our co-designed solution. (a) Training data is dominated by successful trajectories, but real long-horizon execution routinely involves mistakes and detours, leaving recovery and complex-UI behavior weakly supervised. (b) Long browser trajectories often share long prefixes and differ at only a few decisive branch steps, so uniform trajectory-level optimization dilutes the learning signal. (c) Existing benchmarks underrepresent bilingual long-horizon real-web tasks, leaving the deployment regime we target insufficiently evaluated.
}
    \label{fig:challenges}
\end{figure*}

Browser agents built on large language models are moving from curated demonstrations toward real deployment, where completing a task means sustaining dozens of decisions against live, changing web pages: aggregating information across multiple sites, filling multi-stage forms with dependent controls, and coordinating cross-tab workflows. Existing training pipelines and benchmarks, however, remain concentrated on short, successful, English-centric interactions. To quantify this mismatch, we construct BrowserBench, a bilingual benchmark of 350 Chinese--English real-web tasks with an
average completion length of 37.9 steps. The results are sobering: even the
strongest proprietary agent we evaluate fails on roughly one-third of these
tasks, and a strong supervised model succeeds on fewer than one in seven
tasks whose trajectories exceed 50 steps (Section~\ref{sec:difficulty_analysis}). Analyzing these failures, we identify three structural challenges that interlock at the long-horizon regime, illustrated in Figure~\ref{fig:challenges}.

The first challenge is that browser-agent training data is dominated by successful trajectories exactly in the regime where mistakes become routine (Figure~\ref{fig:challenges}(a)). Over a long-horizon real-web task, an agent inevitably encounters unexpected redirects, transient page changes, and complex UI widgets, and must recognize when execution has gone off course and recover toward the goal. Yet training data built only from successful demonstrations provides little supervision on what to do after an error has already occurred. In our analysis, a base supervised model recovers from only 8.5\% of detected erroneous steps, and complex controls such as date pickers, cascaders, and rich-text editors appear too sparsely in naturally collected data for scale alone to teach reliable interaction strategies. The gap is therefore not merely a matter of more data, but of missing supervision on recovery and complex UI behavior.

The second challenge follows from the first: once a browser agent makes a mistake and fails to recover promptly, the trajectory becomes longer, while the supervision signal remains confined to the final task outcome (Figure~\ref{fig:challenges}(b)). In long browser tasks, many rollouts for the same objective share extended prefixes and differ only at a small number of behaviorally decisive branch steps, so treating the entire trajectory uniformly spreads the learning signal over long shared segments instead of concentrating it on the decisions that actually determine success or failure. The problem is further amplified by the nature of browser state itself: navigation invalidates earlier DOM snapshots, local page changes are better captured as compact updates, and optimizing all steps under a single ever-growing transcript forces the policy to reason over obsolete state while paying the highest token cost exactly where trajectories are longest.

The third challenge is that the evaluation landscape still underrepresents the real-world regime that browser agents must handle in deployment (Figure~\ref{fig:challenges}(c)). Existing online benchmarks are overwhelmingly English-centric and often emphasize relatively short tasks, leaving long-horizon interactions on diverse real websites insufficiently tested. As a result, even strong agents may appear competitive under current evaluations while failing on the longer, more compositional workflows that dominate realistic use. The Chinese web, despite its scale and deployment relevance, remains almost entirely absent from prior benchmark construction. Without an evaluation suite that explicitly targets long-horizon, bilingual, real-web tasks, progress on robust browser agents cannot be measured reliably in the setting that matters most.

We address these challenges with a unified pipeline in which the execution substrate, supervision, optimization, and evaluation are co-designed for the same long-horizon deployment setting. Underpinning all stages is a structured browser harness layer that exposes a validated browser action interface and maintains efficient decision-oriented execution contexts, providing an identical interface across supervised data construction, online rollouts, and evaluation. To address the first challenge, we propose Reflection and UI-specialized Curriculum SFT (RUIC-SFT), which augments general demonstrations with reflection-rich recovery trajectories and systematically collected complex-UI interaction data under a three-phase curriculum that stabilizes basic operations before strengthening UI interaction and, finally, self-correction. To address the second, we develop Divergence-Aware Online GRPO (DAO-GRPO), a customized online reinforcement learning framework that improves long-horizon optimization under sparse browser feedback by introducing denser progress supervision, emphasizing behaviorally critical decisions, and aligning training with the step-wise decision contexts encountered during execution. To address the third, we construct BrowserBench, a long-horizon bilingual benchmark over real Chinese and English websites, with goal-only instructions and realistic multi-step tasks designed to reflect deployment conditions more faithfully.

Experiments on WebVoyager, Online-Mind2Web, and BrowserBench show that the proposed pipeline yields consistent and interpretable gains. RUIC-SFT improves substantially over positive-only supervision, especially on recovery-related and complex-UI tasks, while DAO-GRPO adds further gains that grow with task difficulty and trajectory length, consistent with its long-horizon design. BrowserBench further exposes capability differences that are obscured by aggregate scores alone, and qualitative analysis shows the resulting agents detecting erroneous navigation from environment feedback, revising page hypotheses, and recovering within a single episode. At the 27B scale, Wuying-Browser-Agent establishes a new open-source state of the art on browser-use benchmarks, and the same harness-grounded training also transfers to broader tool-use benchmarks including Tau2-Bench, BFCL-v4, and Claw-Eval.

Our contributions are as follows:

\begin{itemize}
\item We identify three interlocking structural challenges in long-horizon browser-agent deployment: training data dominated by successful trajectories but lacking recovery supervision, long trajectories where final-outcome feedback fails to highlight the few decisions that determine success, and an evaluation landscape that underrepresents bilingual long-horizon real-web tasks. We address these challenges jointly through a co-designed pipeline.

\item We build a structured browser harness layer, a validated tool space with decision-oriented task-level state management, that serves as the shared execution substrate across supervised training, online reinforcement learning, and evaluation.

\item We propose RUIC-SFT, a curriculum-based supervised initialization that combines reflection-rich recovery supervision with UI-specialized interaction data under a progressive mixing schedule.

\item We develop DAO-GRPO, a customized online reinforcement learning framework for long-horizon browser tasks that improves optimization under sparse feedback and concentrates learning on behaviorally important decisions.

\item We construct BrowserBench, a long-horizon bilingual evaluation suite of 350 Chinese--English real-web tasks averaging 37.9 steps, addressing the lack of realistic long-horizon online benchmarks beyond the current English-centric setting.

\item We release Wuying-Browser-Agent, a series of open-source browser agents at the 4B, 9B, and 27B scales that establishes a new open-source state of the art on browser benchmarks while retaining strong general tool-use capability.
\end{itemize}

\section{Related Work}

\subsection{Browser Agents}
\label{sec:related_agents}

Recent advances in browser agents have been driven by stronger foundation models, grounding strategies, and web-specific post-training. General-purpose LLMs~\cite{yang2025qwen3, xu2026deepseek} and VLMs~\cite{bai2025qwen3, wang2025internvl3} now provide the backbone for modern browser agents. On the framework side, ReAct~\cite{yao2022react} has established the reasoning-and-acting paradigm that most browser agents follow, while SeeAct~\cite{zheng2024gpt} demonstrates that GPT-4V can serve as a generalist web agent when combined with structured grounding, and AutoWebGLM~\cite{lai2024autowebglm} bootstraps a web navigation agent through automated data collection and reinforcement.

A growing body of work trains agents specifically for browser and GUI interaction. UI-TARS~\cite{qin2025ui} and UI-TARS-2~\cite{wang2025ui} develop native GUI agent models that perceive screenshots directly, while FARA~\cite{awadallah2025fara}, MolmoWeb~\cite{gupta2026molmoweb} and ScaleCUA~\cite{liu2025scalecua} demonstrate that compact models with efficient agentic design or curated demonstrations can achieve strong performance. OpenWebVoyager~\cite{he2025openwebvoyager} explores iterative real-world exploration and feedback for building multimodal web agents. Evaluation has broadened from self-hosted environments~\cite{zhou2024webarena} and static demonstrations~\cite{deng2023mind2web} toward live-web benchmarks including WebVoyager~\cite{he2024webvoyager}, Online-Mind2Web~\cite{xue2025illusion}, VisualWebArena~\cite{koh2024visualwebarena}, and DeepShop~\cite{lyu2025deepshop}. Despite this rapid progress, most existing systems focus on model architecture or training algorithms in isolation, without jointly addressing recovery-oriented data construction, stable long-horizon online optimization, and evaluation in the bilingual real-web setting. Wuying-Browser-Agent bridges this gap through an end-to-end framework where BrowserBench identifies capability-specific weaknesses, RUIC-SFT provides structured SFT initialization, and DAO-GRPO further refines the policy through divergence-aware online reinforcement learning.

\subsection{Computer-Use and Generalist Browser Agents}
\label{sec:related_cua}

Beyond browser-specific systems, a parallel line of work develops
computer-use agents (CUAs) trained to control general-purpose GUI
environments across desktop, web, and mobile platforms, several of
which serve as the open-source baselines in our experiments.
OpenCUA~\cite{wang2026opencua} scales computer-use supervision with
large volumes of OS-level interaction data to obtain strong
generalist GUI control, while
GUI-Owl-1.5~\cite{xu2026mobile} builds unified vision-language
agents for cross-platform GUI understanding and action grounding.
The UI-Venus family~\cite{team2026ui} pursues compact
mixture-of-experts GUI agents, and
EvoCUA~\cite{huang2026evocua} shows that multi-turn online RL
significantly strengthens computer-use agents' general tool-use
capability. In the same spirit, Qwen-UI-Agent~\cite{zhou2026qwen}
provides a general-purpose agent grounded in UI interaction, and
OpenWebRL~\cite{yang2026openwebrl} demonstrates that compact open
models become competitive through online multimodal RL on live web
pages. 

However, most CUA systems are optimized for atomic GUI control or
general computer-use ability, and are typically evaluated under
relatively short-horizon or single-platform settings. They do not
jointly address the three long-horizon challenges targeted in this
work: recovery-oriented supervision for error-prone real-web
execution, branch-sensitive credit assignment under sparse terminal
rewards, and bilingual long-horizon evaluation. Wuying-Browser-Agent
complements this line by grounding the policy in a validated
browser harness and optimizing it specifically for sustained,
real-world web interaction, while retaining competitive general
agentic capability.

\subsection{Agentic Reinforcement Learning}
\label{sec:related_rl}

Recent success in outcome-based reinforcement learning for language reasoning~\cite{guo2025deepseek} has accelerated the use of RL for interactive agents. Methods such as GRPO~\cite{shao2024deepseekmath} and DAPO~\cite{yu2026dapo} enable critic-free policy optimization through group-relative advantages, while Visual-RFT~\cite{liu2025visual}, VLM-R1~\cite{shen2025vlm}, and UI-R1~\cite{lu2026ui} extend reinforcement fine-tuning to multimodal models for visual reasoning, referring expression comprehension, detection and GUI tasks. A two-stage paradigm has therefore become common: supervised fine-tuning provides a stable initialization, and reinforcement learning further improves the policy through task-driven optimization.

For browser agents, static supervision alone is insufficient. Browser tasks are interactive, stateful, and path-dependent, so capabilities such as long-horizon exploration, recovery from intermediate failure, and adaptation to unseen page variations cannot be learned reliably from fixed demonstrations alone. This makes online RL particularly important. WebRL~\cite{qi2025webrl} studies curriculum-based online optimization in WebArena, while AgentRL~\cite{zhang2025agentrl}, WebAgent-R1~\cite{wei2025webagent}, and RAGEN~\cite{wang2025ragen} explore multi-turn GRPO-style training and self-evolution in simulated or self-hosted web environments. Moving toward more realistic settings, PAE~\cite{zhou2025proposer} focuses on autonomous skill discovery, WebGym~\cite{bai2026webgym} provides scalable open-web training environments, and OpenWebRL~\cite{yang2026openwebrl} shows that compact 4B models can become competitive through effective warm-starting and online multimodal GRPO. 

However, stable online optimization for browser agents remains difficult. One challenge is long-horizon credit assignment under sparse rewards: many rollout steps are shared prefixes that carry little discriminative signal. Another is that browser contexts are not append-only: DOM snapshots may be inserted, replaced, or deleted across navigation steps, so standard sequence-level objectives do not match the information actually available at decision time. Prior work addresses parts of this problem through dense shaping or intermediate supervision. Potential-based reward shaping (PBRS)~\cite{ng1999policy} provides dense rewards while preserving the optimal policy, and process reward models score intermediate steps at substantial annotation cost. DAO-GRPO instead combines PBRS-style dense supervision with an LLM-based divergence detector for branch-sensitive credit assignment, and adopts a response-level objective tailored to dynamically reconstructed browser contexts.

\subsection{Browser Agent Benchmarks}
\label{sec:related_eval}

Existing browser-agent benchmarks differ in realism, coverage, and evaluation protocol. WebArena~\cite{zhou2024webarena} provides self-hosted tasks with deterministic success criteria, while Mind2Web~\cite{deng2023mind2web} contributes large-scale demonstrations but relies on offline HTML snapshots. WebVoyager~\cite{he2024webvoyager} moves to real websites with LLM-based judgment, and Online-Mind2Web~\cite{xue2025illusion} highlights the impact of temporal website changes. More recent benchmarks target specific dimensions, such as long-horizon information gathering~\cite{gou2026mind2web}, shopping~\cite{lyu2025deepshop}, and hard information retrieval~\cite{wei2025browsecomp}.

Despite this progress, three limitations remain. First, long-horizon tasks are still underrepresented. Second, existing benchmarks are overwhelmingly English-centric, leaving the Chinese web almost entirely unevaluated. Third, website coverage is often narrow; for example, WebVoyager contains 643 tasks from only 15 websites. As a result, current evaluation does not adequately reflect the bilingual, long-horizon, real-web setting targeted in this work.

We address these limitations with \textbf{BrowserBench}, a bilingual long-horizon browser-agent benchmark of 350 real-web tasks averaging 37.9 steps and spanning 254 websites. Each task is normalized into a goal-only instruction paired with a structured success criterion, enabling reliable automated scoring. Cases are further annotated by difficulty and task pattern for capability-level diagnosis beyond a single aggregate score. BrowserBench complements existing benchmarks by extending rigorous evaluation to the long-horizon, bilingual real-web regime so far absent from the literature.

\section{Preliminaries}\label{sec:preliminaries}

\subsection{Problem Formulation}
\label{sec:problem_formulation}

We formulate browser-agent execution as a partially observable sequential decision-making problem. Given a natural-language task instruction $g$ and an initial browser state, the agent interacts with the browser environment for at most $T_{\max}$ steps. At each step $t$, the environment provides an observation $o_t$, the agent produces a browser response $y_t$, and the controller executes a structured action $a_t$ parsed from $y_t$. The environment then returns feedback $e_t$ and transitions to the next browser state.

\paragraph{Observation space.}
The primary observation is a structured browser-state representation derived from the current page DOM, denoted by $S_t$. This representation preserves interactive elements and relevant structural information while removing non-visible or irrelevant content. When DOM information alone is insufficient for grounding---for example, when visual layout, rendered content, or image-based cues are required---the agent additionally receives a viewport screenshot $V_t$. We therefore define the observation as
\begin{equation}
    o_t =
    \begin{cases}
        (S_t, V_t), & \text{if visual grounding is required},\\
        S_t, & \text{otherwise}.
    \end{cases}
    \label{eq:observation}
\end{equation}

\paragraph{Decision context.}
The policy does not operate on the current observation alone. Instead, at each step it conditions on a \emph{decision context} $c_t$, reconstructed from the task instruction, the current observation, previous actions, and structured environment feedback. Formally, we write
\begin{equation}
    c_t = \mathcal{R}\!\left(g,\; o_t,\; \{a_u, e_u\}_{u < t}\right),
    \label{eq:context_reconstruct}
\end{equation}
where $\mathcal{R}(\cdot)$ denotes the browser-context reconstruction operator. Thus, the policy input at step $t$ may contain: (i) the task instruction, (ii) the current structured browser state $S_t$, (iii) an optional screenshot $V_t$ when visual grounding is needed, and (iv) retained interaction history in the form of prior actions and environment feedback. This distinction is important because browser-agent contexts are inherently dynamic rather than simple append-only transcripts. Effective execution therefore requires a context management mechanism that keeps the information relevant to the current decision while avoiding unnecessary redundancy.

\paragraph{Action space.}
The agent acts through a unified structured browser action space spanning navigation, interaction, extraction, file operations, and flow control (detailed in Section~\ref{sec:harness}). Each action is parameterized by target elements or values grounded in the current observation. The model generates a browser response $y_t$, which may contain structured action content and optional intermediate reasoning; the harness then deterministically parses $y_t$ into an executable action $a_t$ in the predefined action vocabulary.

\paragraph{Trajectory and objective.}
A trajectory is written as
\begin{equation}
    \tau = \{(o_t, c_t, y_t, a_t, e_t)\}_{t=1}^{T},
    \label{eq:trajectory}
\end{equation}
where $T \le T_{\max}$ is the termination step. Given task instruction $g$, the objective is to produce a trajectory whose terminal browser state satisfies the task specification. Task success is determined by an external judge module operating on the final state and interaction log. During execution, the agent does not directly observe task reward and must infer progress from browser observations and environment feedback.

\subsection{Browser Environment Interface}
\label{sec:agentbay}

All browser interactions in this work are executed in Wuying AgentBay~\cite{piao2025agentbay}, a cloud-native sandbox service that provides secure and isolated browser environments for autonomous agent execution. Each task is assigned to an independent browser sandbox hosted in a standardized Linux environment, ensuring consistent execution across parallel runs.

The agent communicates with the sandbox through a lightweight Model Context Protocol (MCP) interface that exposes compact browser primitives for session initialization, state acquisition, action execution, metadata retrieval, and structured output control. In particular, action execution and state-transition retrieval are coupled within a unified environment call, which avoids repeatedly transmitting full browser states and substantially reduces interaction overhead during rollout. This property is especially important for online training, where browser trajectories may contain dozens of decision steps and environment communication can otherwise become a major bottleneck. Additional implementation details are provided in the experimental setup.

\begin{figure}
    \centering
    \includegraphics[width=0.98\textwidth]{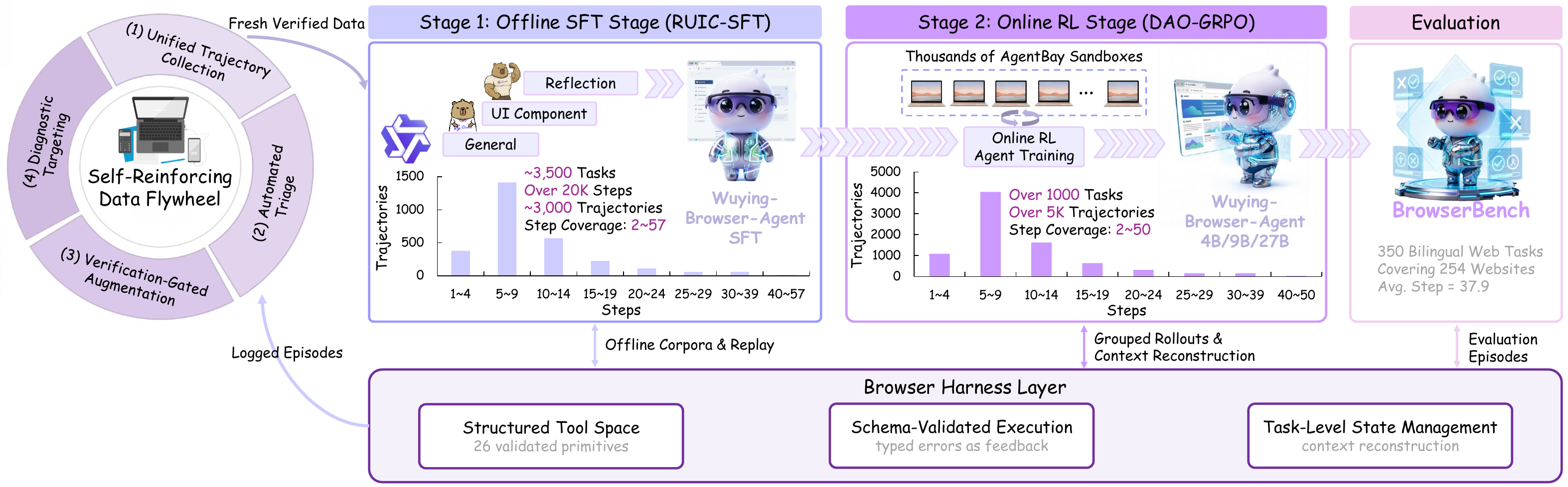}
    \caption{
Overview of the Wuying-Browser-Agent pipeline. A shared browser harness layer supports two-stage policy training, with RUIC-SFT learning from around 3K curated trajectories and DAO-GRPO refining the policy through over 5K live-web rollouts. BrowserBench provides bilingual long-horizon evaluation, while a self-reinforcing data flywheel feeds verified interaction outcomes back into subsequent training.
}
    \label{fig:network_overview}
\end{figure}

\section{Method}\label{sec:method}

Our method is designed to close the robustness gap identified in the introduction: browser agents trained mainly on successful trajectories remain brittle when they must recover from mistakes, interact with complex UI controls, and make correct decisions over long, changing web interactions. Figure~\ref{fig:network_overview} provides an overview of the full Wuying-Browser-Agent pipeline. At its core, the policy is trained in two stages. The first stage, Reflection and UI-Specialized Curriculum SFT (RUIC-SFT), provides a robustness-oriented supervised initialization. The second stage, Divergence-Aware Online GRPO (DAO-GRPO), further improves the same policy through online interaction, focusing optimization on branch-defining decisions under dynamically reconstructed browser contexts.

Both stages, together with evaluation, run on the browser harness layer described in Section~\ref{sec:harness}, which provides the structured tool space, schema-validated execution, and decision-oriented context management shared across offline supervised learning and online policy refinement. To reduce observation noise and token cost, the harness applies an optimized DOM simplification pipeline that preserves decision-relevant interactive elements instead of serializing full raw DOM trees. We first present the harness layer, then describe RUIC-SFT and DAO-GRPO.

\begin{table}[t]
\centering
\caption{\textbf{Browser Agent Action Space.} We define a structured browser action space with 24 atomic operations used in both SFT and online RL training. Actions are organized into five functional categories, including navigation, interaction, extraction, file operations, and flow control.}
\label{tab:action_space}
\small
\setlength{\tabcolsep}{5pt}
\begin{tabular}{@{}ll@{}}
\toprule
\textbf{Category} & \textbf{Representative Actions} \\
\midrule
\textbf{Navigation} 
& \texttt{go\_to\_url}, \texttt{search}, \texttt{go\_back}, \texttt{switch\_tab}, \texttt{close\_tab} \\

\textbf{Interaction} 
& \texttt{click}, \texttt{hover}, \texttt{input}, \texttt{scroll}, \texttt{select\_dropdown}, \texttt{drag}, \texttt{set\_slider}, \texttt{upload}, \texttt{send\_keys} \\

\textbf{Extraction} 
& \texttt{extract}, \texttt{screenshot}, \texttt{save\_pdf}, \texttt{eval\_javascript} \\

\textbf{File Operations} 
& \texttt{read\_file}, \texttt{write\_file} \\

\textbf{Flow Control} 
& \texttt{update\_plan}, \texttt{load\_skill}, \texttt{wait}, \texttt{done} \\
\bottomrule
\end{tabular}
\end{table}

\subsection{Browser Harness Layer}\label{sec:harness}

Long-horizon browser interaction requires more than a strong policy: model outputs must be parsed, validated, and executed reliably against live pages, and the context fed back to the model must remain faithful to the current task state. We therefore build a browser harness layer above the underlying policy model. The harness serves as the execution substrate that connects model outputs to the browser environment, and it is shared by both RUIC-SFT and DAO-GRPO.

\subsubsection{Structured Tool Space}
The harness exposes a structured browser action space with 24 atomic operations, summarized in Table~\ref{tab:action_space}. These actions span five functional categories, including navigation, interaction, extraction, file handling, and flow control. At each interaction step, the model emits a structured action over this tool space as its browser response $y_t$.

\subsubsection{Structured Execution and Feedback}
The harness parses each emitted call, validates its schema and parameters against the tool definitions, dispatches it to the corresponding executor in the AgentBay sandbox (Section~\ref{sec:agentbay}), and returns structured feedback $e_t$ together with the updated browser observation $o_{t+1}$. Invalid calls are rejected with typed error messages that re-enter the context as feedback, allowing the policy to detect and correct its own malformed outputs rather than failing silently. This establishes a tightly coupled interaction process in which perception, decision, execution, and state update remain linked at every step.

\subsubsection{Task-Level State Management}
A key function of the harness is task-level state management. Rather than exposing the model to an ever-growing raw interaction transcript, the harness maintains an efficient decision-oriented context that keeps the most relevant task state, action history, and environment feedback available at each step. It also manages intermediate artifacts such as extracted content and temporary files, enabling cross-step memory and result delivery for tasks whose outcome is a file or an aggregated report. Because this shared harness abstraction is used consistently in RUIC-SFT data construction, online RL rollouts, and evaluation, the policy experiences an identical interface across offline learning, online interaction, and deployment, allowing direct transfer between stages. More broadly, this design trains the policy to operate over a schema-constrained tool interface with explicit state tracking and feedback grounding, which helps preserve transfer to more general agentic settings beyond browser-specific benchmarks.

\begin{figure}
    \centering
    \includegraphics[width=0.98\textwidth]{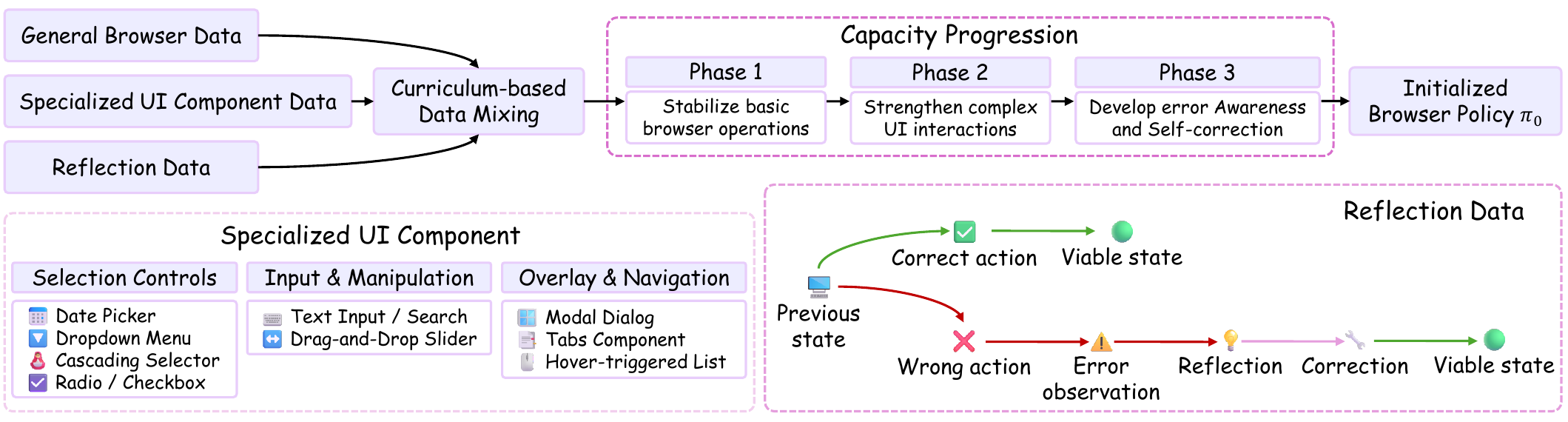}
    \caption{
    Piecewise linear annealing schedule for dataset mixing in RUIC-SFT. Training begins with a general-data-dominant distribution to stabilize basic browser operations, gradually increases the proportion of specialized UI component data to strengthen complex interaction skills, and introduces reflection data only in the late stage to cultivate self-correction without destabilizing the execution prior.
    }
    \label{fig:rui_csft_overview}
\end{figure}

\subsection{Reflection and UI-specialized Curriculum SFT (RUIC-SFT)}\label{sec:RUIC-SFT}

The goal of the supervised stage is not merely to imitate successful browser behavior, but to provide a robustness-oriented initialization for realistic web interaction. In particular, we target two forms of supervision that are systematically missing from conventional positive-only browser demonstrations: recovery after off-trajectory errors and reliable interaction with underrepresented complex UI controls. To address these gaps, we construct two specialized data sources in addition to general browser task data: a reflection dataset for error detection and recovery, and a specialized UI dataset for complex control manipulation. We integrate them through \textbf{Reflection and UI-specialized Curriculum SFT (RUIC-SFT)}.

\subsubsection{Why Successful Demonstrations Alone Are Insufficient}

Simply increasing the volume of generic browser trajectories~\cite{zhou2024webarena,deng2023mind2web} is unlikely to resolve either deficiency. General successful demonstrations teach the model how to act correctly when execution remains on the intended path, but provide little supervision on what should happen after the model has already deviated from that path. Likewise, complex UI widgets appear only sparsely in naturally collected browser trajectories, with heavily skewed type distributions, making it difficult for scale alone to induce robust interaction policies. In other words, the missing supervision is structural rather than merely quantitative. The reflection dataset explicitly exposes the model to trajectories with an error--awareness--reflection--correction structure, while the specialized UI dataset systematically concentrates training on interaction patterns that are under-represented in general web data.

From a learning perspective, the two datasets provide complementary supervision. The specialized UI dataset reduces uncertainty in action selection over complex controls by repeatedly exposing the model to structured interaction patterns for difficult widgets. The reflection dataset, in contrast, provides supervision specifically for off-trajectory recovery, which is largely absent from successful demonstrations alone. Together, they improve not only the probability of executing a difficult task correctly on the first attempt, but also the model's ability to recover when deviations still occur.

\subsubsection{Reflection Dataset}
\label{sec:reflection_data}

Each reflection example is organized around an erroneous step $t^*$. In addition to the task context and the incorrect action itself, the example records the resulting error observation, a natural-language reflection explaining why the action was inappropriate under the current page state, and a correction strategy consisting of actions that return the agent to a viable execution path. The dataset covers a broad range of common browser-agent failure patterns, including navigation mistakes, interaction errors, state misinterpretation, and ineffective repeated behavior. All correction strategies are executed in a sandbox environment to verify that they genuinely recover the task.

Reflection examples are collected through a combination of automated mining and human-in-the-loop curation. Failed trajectories generated during bootstrapping are analyzed to localize likely error segments, after which annotators provide reflection reasoning and corrective continuations. We further expand coverage by constructing additional recovery examples around underrepresented failure patterns and by preserving matched successful and unsuccessful interaction traces when useful for future preference-based training.

Data quality is controlled through a combination of execution verification, annotation review, and distributional monitoring. In particular, reflection reasoning must be grounded in concrete state transitions, and correction sequences must be executable and demonstrably useful in restoring a viable task path.

As a result, the reflection dataset does not merely expose the model to failed trajectories; it teaches the policy to treat unexpected observations as evidence of possible off-path execution and to generate grounded corrective continuations, which are central to robust browser deployment.

\subsubsection{Specialized UI Component Dataset}
\label{sec:ui_data}

Modern web applications widely employ complex UI controls whose interaction patterns differ substantially from those of ordinary text inputs and buttons. Examples include date pickers that require hierarchical year--month--day selection, cascaders that rely on progressive expansion and linked confirmation, and rich-text editors involving multi-step toolbar interactions. These controls are often implemented differently across frontend frameworks and websites, so sparse natural exposure is insufficient for the model to internalize reliable interaction strategies. This creates a systematic capability bottleneck in browser-agent deployment.

To address this issue, we construct a specialized UI component dataset through a four-stage pipeline. We first collect component scenes from real websites and establish mappings among component type, website, page, and concrete control instance using a combination of automated DOM-structure scanning and manual verification. We then design parameterized interaction tasks for each control type with progressively increasing difficulty, ranging from simple single-step selection tasks to compound operations and edge cases involving invalid inputs or exceptional control states. Next, human annotators operate the real controls and record action sequences, key operation nodes, and control state transitions such as panel expansion, highlighting changes, and linked updates. Finally, we augment the dataset through cross-website transfer of interaction strategies, generation of equivalent action-sequence variants, and injection of anomalous states such as loading delays, disabled controls, and validation failures. These anomalous UI states also create a natural bridge to the reflection dataset, allowing the two supervision sources to reinforce one another.

This concentrated supervision reduces the reliance on sparse natural exposure and equips the policy with reusable interaction patterns for UI structures that frequently trigger failures in real deployment.

\subsubsection{Curriculum-Based Training Schedule}
\label{sec:curriculum}

Figure~\ref{fig:rui_csft_overview} provides an overview of the three data sources and their curriculum-based integration. Rather than training sequentially on each data source or using a fixed mixture from the beginning, we adopt a curriculum-based mixed training strategy with progressive ratio annealing~\cite{bengio2009curriculum}. The capability dependencies in browser-agent learning suggest a natural order: the model should first stabilize basic browser actions, then strengthen its ability to manipulate specialized UI components, and only afterward emphasize self-correction behavior. At the same time, purely sequential training risks catastrophic forgetting, especially when late-stage reflection-heavy training shifts the model toward overly cautious behavior. We therefore combine curriculum learning with progressive mixture control.

Let $D_g$, $D_u$, and $D_r$ denote the general browser task dataset, the specialized UI component dataset, and the reflection dataset, respectively. Rather than using a fixed mixture throughout training, we adopt a dynamic mixing strategy that changes over the course of training. As illustrated in Figure~\ref{fig:rui_csft_overview}, training begins with a general-data-dominant distribution to stabilize basic browser operations, gradually increases the proportion of specialized UI data to strengthen complex interaction skills, and introduces reflection data only in the later stage to cultivate self-correction without destabilizing the execution prior. The transition schedule is selected based on validation behavior.

The three phases serve distinct purposes. \textbf{Phase~1} is dominated by general browser trajectories, with a small amount of UI-specialized exposure, and is primarily used to stabilize foundational capabilities such as navigation, element grounding, clicking, typing, and action formatting. \textbf{Phase~2} gradually shifts emphasis toward complex component interaction while retaining sufficient general data to preserve the browser-operation prior acquired in Phase~1. \textbf{Phase~3} introduces reflection data while
annealing the UI-specialized proportion to a lower level. This delayed introduction of $D_r$ is deliberate: preliminary experiments showed that introducing reflection supervision before the execution policy had stabilized led to overly conservative behavior, with the agent becoming excessively prone to backtracking and over-explaining even on straightforward tasks. By introducing reflection only after the execution prior and UI interaction skills have been sufficiently strengthened, we obtain a more balanced model that can both act decisively and recover when necessary. The phase boundaries and endpoint ratios are determined from preliminary tuning; Section~\ref{sec:ablation} compares this schedule against fixed-ratio mixing and an aggressive early-reflection variant.

\subsubsection{Training Objective}

Under this schedule, the learning objective remains the standard supervised next-token prediction loss, with the effective training distribution changing over time:
\begin{equation}
    \mathcal{L}_{\mathrm{SFT}}(\theta)
    =
    \mathbb{E}_{(x,y)\sim p_g(\lambda)D_g + p_u(\lambda)D_u + p_r(\lambda)D_r}
    \left[
        -\log P_{\theta}(y \mid x)
    \right].
    \label{eq:sft_loss}
\end{equation}
The curriculum is therefore realized through dynamic sampling rather than by changing the optimization objective itself. This allows us to retain the optimization stability of conventional SFT while altering the effective supervision distribution in a capability-aware manner.

From a functional perspective, the three data sources contribute distinct but complementary signals. General data $D_g$ provides the foundational prior for standard browser operation. Specialized UI data $D_u$ improves recognition and manipulation of complex controls whose interaction logic cannot be reliably inferred from sparse natural coverage. Reflection data $D_r$ teaches the model to interpret unexpected observations as evidence of possible failure, attribute the error to plausible preceding actions, and generate corrective continuations. Because these capabilities are introduced through a progressively annealed mixture rather than abrupt dataset switching, the resulting model is less prone to forgetting previously acquired behavior and better able to balance direct task execution with recovery behavior.

In summary, RUIC-SFT provides a capability-structured initialization that extends beyond behavioral cloning to establish a structured credit prior for subsequent online optimization. By explicitly annotating step-level error-recovery pairs, the reflection dataset supplies dense pseudo-labels for what would otherwise be sparse terminal rewards in long-horizon tasks. Consequently, DAO-GRPO initializes from a policy with inherent causal attribution capability, allowing online RL to concentrate its optimization budget on refining branch-defining decisions rather than discovering recovery primitives from scratch. This structured handoff motivates the divergence-aware online refinement stage described next.

\begin{figure*}
    \centering
    \includegraphics[width=0.98\textwidth]{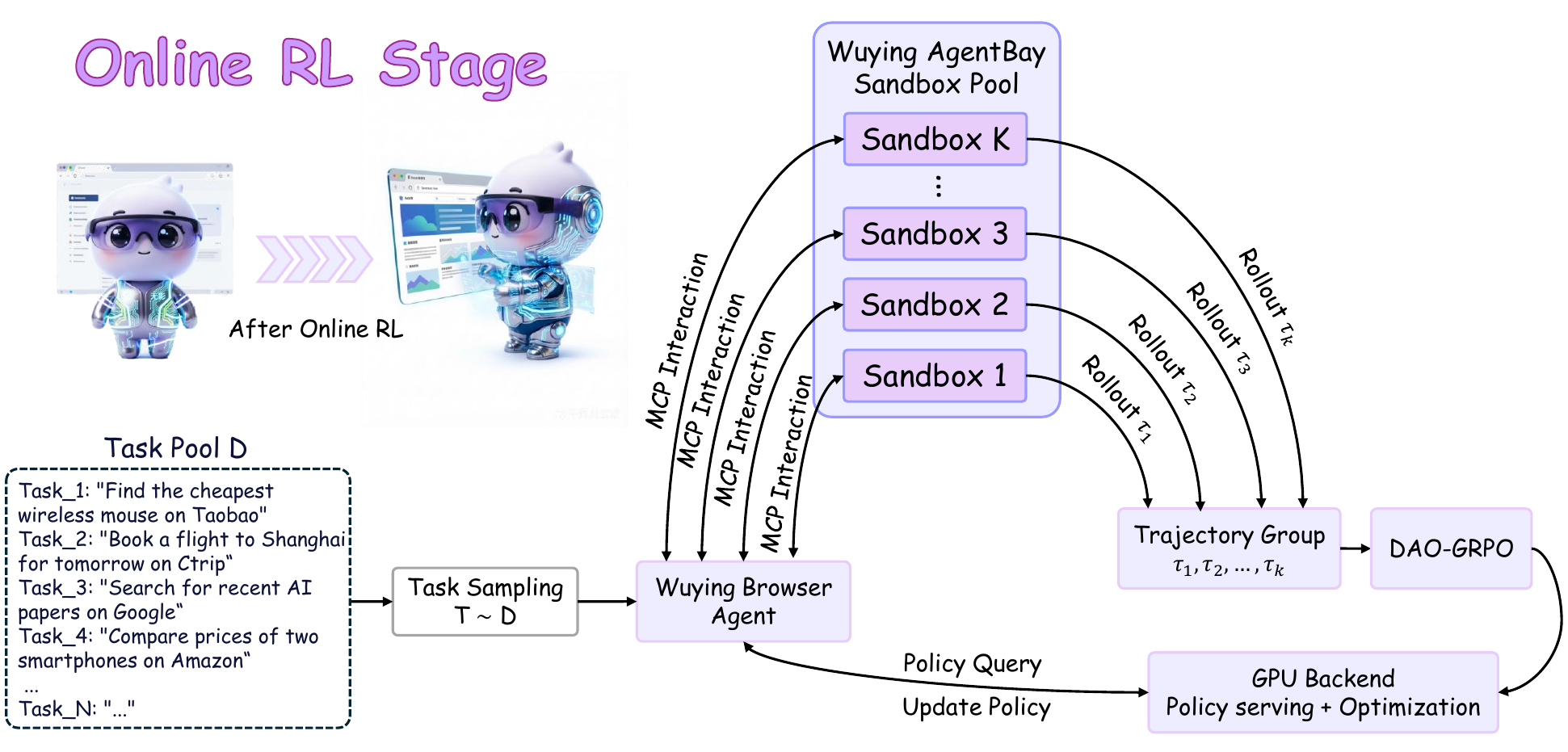}
    \caption{
    System overview of the online reinforcement learning stage. A task is sampled from the task pool $\mathcal{D}$ and assigned to the browser agent controller, which interacts with multiple AgentBay sandboxes in parallel. Each sandbox produces one rollout trajectory, and the resulting trajectory group is consumed by the DAO-GRPO optimization module. The same GPU backend supports both policy serving during rollout and parameter updates during online optimization.
    }
    \label{fig:online_rl_overview}
\end{figure*}

\subsection{Divergence-Aware Online GRPO (DAO-GRPO)}\label{sec:DAO-GRPO}

Starting from the robustness-oriented initialization learned by RUIC-SFT, we further optimize the same browser policy through direct interaction with live browser environments. This online stage is necessary because offline supervision alone cannot fully optimize robust browser behavior: in realistic web tasks, success often depends on a small number of branch-defining decisions made under sparse feedback and dynamically changing contexts. We therefore develop \textbf{Divergence-Aware Online GRPO (DAO-GRPO)}, an online policy optimization framework designed to refine long-horizon robustness under real browser interaction.

Our online RL stage is designed to address three structural difficulties in browser learning: sparse terminal rewards, long trajectories in which only a few decisions are truly outcome-defining, and the dynamic nature of browser context across interaction steps. Rather than relying on a generic trajectory-level objective, we use a browser-tailored optimization strategy that provides denser supervision, emphasizes critical decisions, and better matches the step-wise decision process encountered during execution.

\subsubsection{Online Optimization Overview}
\label{sec:dao_overview}

DAO-GRPO is built around a simple principle: online learning should optimize the browser policy using the same branching structure and step-specific context under which the agent actually acts. Figure~\ref{fig:online_rl_overview} illustrates the system-level interaction loop of online optimization. For each sampled task $d \sim \mathcal{D}$, the current policy $\pi_\theta$ is served by the browser agent controller and interacts with multiple AgentBay browser-use sandbox instances in parallel. Each sandbox executes one browser rollout, producing a trajectory, and the resulting trajectory group is consumed by the DAO-GRPO optimizer for policy updates. The same lightweight browser interface introduced in Section~\ref{sec:agentbay} is used during rollout, enabling efficient interaction without repeatedly serializing full browser states.

For each sampled task, DAO-GRPO generates a grouped rollout
\[
\{\tau_1,\tau_2,\ldots,\tau_K\},
\]
where each trajectory $\tau_i$ contains a sequence of step-specific decision contexts, policy responses, executable browser actions, and environment feedback. At step $t$, the reconstructed decision context $c_{i,t}$ may include the task instruction, the current structured browser state $S_{i,t}$, an optional screenshot $V_{i,t}$ when visual grounding is required, and the retained interaction history consisting of previous actions and environment feedback. For optimization, DAO-GRPO combines three quantities: a trajectory-level relative advantage $A_i$ estimated within the group, a step-level credit weight $w_{i,t}$ derived from cross-trajectory divergence analysis, and the step-specific decision context $c_{i,t}$ reconstructed by the harness. These are fused into a weighted response-level policy objective.


\subsubsection{Grouped Return Estimation}
\label{sec:reward}

A first challenge in online browser learning is that task-level success provides only sparse supervision, while robust long-horizon behavior depends on intermediate progress signals. We therefore combine terminal judgment with potential-based reward shaping (PBRS), which provably preserves the optimal policy in the discounted infinite-horizon setting~\cite{ng1999policy}; our finite-horizon grouped variant inherits this only as a practical approximation. Empirically, adding shaping improves the terminal success rate itself rather than merely inflating shaped returns (Table~\ref{tab:ablation_daograpo}), indicating that the potential signal guides rather than hijacks learning.

For each trajectory $\tau_i$, we obtain a terminal reward $R_i^{\mathrm{term}} \in \{0,1\}$ indicating whether the terminal browser state satisfies the task specification. To densify this sparse signal, we augment it with a potential-based shaping term $r^{\mathrm{shape}}_{i,t} = \gamma \Phi(s_{i,t+1}) - \Phi(s_{i,t})$, where $\Phi(\cdot)$ is a task-conditioned progress estimator and $\gamma$ is a discount factor. The boundary condition $\Phi(s_{T_i})=0$ ensures that shaping does not alter the terminal success criterion. The resulting trajectory return combines both signals:
\begin{equation}
    R_i = R_i^{\mathrm{term}} + \sum_{t=1}^{T_i-1} r^{\mathrm{shape}}_{i,t}.
    \label{eq:return}
\end{equation}
Since trajectories are generated in groups for the same task, we normalize returns within each rollout group to obtain a trajectory-level relative advantage:
\begin{equation}
    A_i = \frac{R_i - \mu_R}{\sigma_R + \epsilon},
    \label{eq:advantage}
\end{equation}
where $\mu_R$ and $\sigma_R$ are the group mean and standard deviation. This grouped normalization yields a task-matched preference signal that is less sensitive to reward-scale variation across tasks. Groups in which all trajectories receive identical returns are discarded, as they carry no relative optimization signal.

\subsubsection{Divergence-Aware Step Credit Assignment}
\label{sec:credit}

A second challenge is that robust browser execution often hinges on a small number of branch-defining decisions: multiple trajectories may share a long common prefix, yet diverge sharply once the agent chooses an incorrect page, element, or action sequence. Uniform trajectory-level weighting therefore fails to focus learning on the decisions that most strongly determine whether execution stays on path or deviates further.


To address this issue, DAO-GRPO estimates step importance from semantic divergences within each trajectory group. Given the $K$ trajectories sampled for the same task, an LLM-based divergence estimator compares them step by step and identifies branch points at which relatively successful and unsuccessful trajectories first differ in a behaviorally meaningful way. For each detected divergence, the estimator outputs (1) a divergence step index $t^*$, (2) a criticality score $c^* \in [0,1]$, and (3) a partition of trajectories into a favorable branch $\mathcal{B}^+$ and an unfavorable branch $\mathcal{B}^-$.

We emphasize that the divergence estimator is used as a semantic credit prior rather than as an exact oracle. The direction of optimization is still determined by the trajectory-level advantage $A_i$ in Eq.~\eqref{eq:advantage}; divergence estimation only redistributes update magnitude across steps.

For trajectory $\tau_i$ and step $t$, we define the step credit weight as:
\begin{equation}
    w_{i,t} =
    \begin{cases}
        \alpha_{\mathrm{shared}}, & t < t^*,\\[4pt]
        \alpha_{\mathrm{div}} \cdot c^*, & t = t^*,\\[4pt]
        w_{i,t^*}(\delta^+)^{\,t-t^*}, & t > t^* \text{ and } \tau_i \in \mathcal{B}^+,\\[4pt]
        w_{i,t^*}(\delta^-)^{\,t-t^*}, & t > t^* \text{ and } \tau_i \in \mathcal{B}^-,
    \end{cases}
    \label{eq:credit}
\end{equation}
where $\alpha_{\mathrm{shared}}$ and $\alpha_{\mathrm{div}}$ are scaling constants, and $\delta^+,\delta^- \in (0,1)$ are asymmetric decay factors satisfying $\delta^+ > \delta^-$. Shared-prefix steps receive a smaller base weight because they contribute little to distinguishing successful from unsuccessful branches. The divergence step receives an emphasized weight modulated by the estimated criticality score, and post-divergence steps on favorable branches retain credit longer than those on unfavorable branches. The specific values of these hyperparameters are tuned on the validation set and omitted here to encourage adaptation to different browser environments.

In practice, the resulting step weights are normalized to maintain stable optimization across trajectories of different lengths. Overall, this mechanism acts as a robustness-oriented credit prior: it does not change which trajectories are preferred globally, but sharpens where the learning signal is concentrated within them.

\subsubsection{Response-Level Policy Optimization}
\label{sec:response_level}

A third challenge is a train--test mismatch in browser-policy optimization. During real interaction, the agent acts under a step-specific context that is dynamically reconstructed from the current page state, retained history, and optional visual evidence. If training instead optimizes all responses under a single concatenated trajectory transcript, the policy is updated under contexts that differ from those actually available at decision time.

Unlike standard GRPO, which assumes an append-only context and optimizes over full trajectory likelihoods, DAO-GRPO decouples context reconstruction from advantage estimation: the group-relative advantage $A_i$ retains GRPO's variance-reduction benefit, while each $\log\pi_\theta(y_{i,t}\mid c_{i,t})$ is
evaluated under the exact information set available at decision time. We perform a single on-policy update per rollout batch, so no importance correction is required.

The trajectory-level preference signal and the step-level credit signal are combined into a weighted training target for each step. In this way, optimization is concentrated on semantically decisive branch steps rather than diluted across long shared prefixes.

Each step-specific context $c_{i,t}$ is produced by the harness state manager (Section~\ref{sec:harness}): DAO-GRPO replays the interaction log through the same reconstruction rules used at online inference, so that $\log\pi_\theta(y_{i,t}\mid c_{i,t})$ is evaluated under exactly the information set available at decision time.

In browser interaction, the information available to the policy at each step is determined by the current page state together with the relevant retained history, rather than by a naively concatenated transcript of everything that has happened before. Training should therefore follow the same context organization used at inference time. Otherwise, the policy may be optimized under contexts that are unnecessarily verbose, partially outdated, or inconsistent with the actual online decision process.

For this reason, our online RL stage optimizes each model response under the step-specific decision context actually available at that point in the interaction, while regularizing the policy against excessive drift from the supervised initialization. This can be interpreted as an advantage-weighted policy update under dynamically managed browser contexts, rather than imitation over a single fixed trajectory transcript.

In practice, each $c_{i,t}$ is reconstructed by replaying the interaction log up to step $t$ using the same state replacement and diff-appending rules as in online inference. When visual grounding is unnecessary, the screenshot block is simply omitted from the reconstructed context. As a consequence, non-critical steps receive only weak update strength, while semantically decisive branch steps dominate the optimization signal.

The updated policy is then used in the next round of grouped parallel rollout. This design aligns optimization more closely with the actual online inference condition, reducing context mismatch and improving robustness to changing browser states.

\section{BrowserBench: A Bilingual Long-Horizon Real-Web Benchmark}
\label{sec:BrowserBench}

\begin{figure*}[t]
    \centering
    \includegraphics[width=0.99\textwidth]{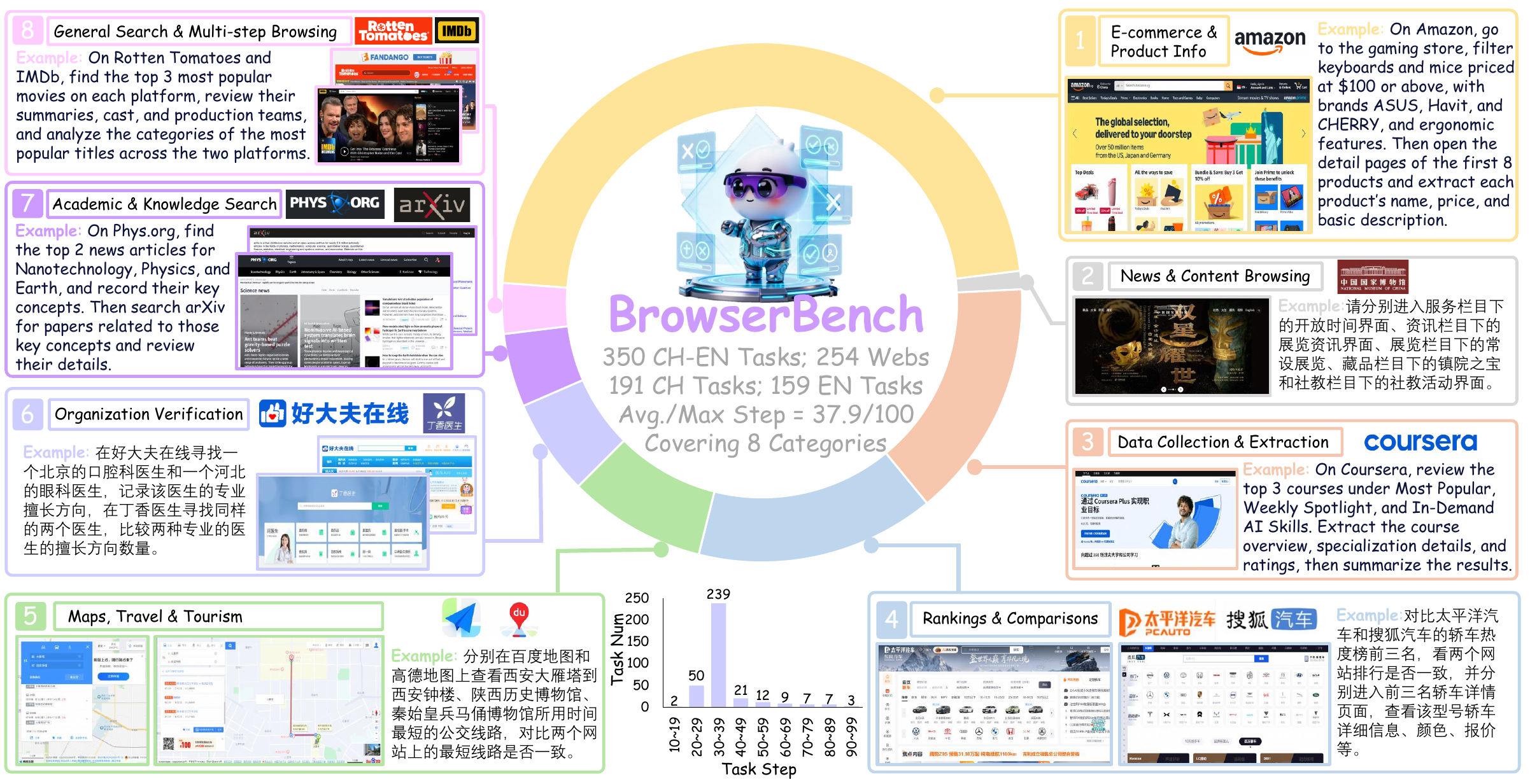}
    \caption{
    Overview of BrowserBench. BrowserBench contains 350 bilingual real-web tasks spanning 254 websites, with an average completion length of 37.9 steps. The benchmark covers eight task categories across Chinese and English websites, and each case is normalized into a goal-only instruction paired with a structured success criterion, enabling reliable Pass@1 evaluation of long-horizon browser agents.
    }
    \label{fig:browserbench_overview}
\end{figure*}

To evaluate browser agents in the deployment regime targeted by this work, we construct BrowserBench, a bilingual long-horizon real-web benchmark. Existing browser-agent benchmarks are still predominantly English-centric and often concentrate on relatively short interactions, leaving sustained multi-step browsing over realistic Chinese--English websites insufficiently evaluated. BrowserBench is designed to fill this gap, and an overview is shown in Figure~\ref{fig:browserbench_overview}. It contains 350 real-web tasks spanning 254 websites in Chinese and English.

The benchmark covers a broad range of realistic domains, including e-commerce, rankings and comparisons, data collection, maps and travel, organization verification, academic search, general search, and news browsing. In language distribution, BrowserBench includes 191 Chinese tasks (54.6\%) and 159 English tasks (45.4\%). In horizon, it is explicitly long-horizon: the average completion length is 37.9 interaction steps, with a minimum of 15 and a maximum of 100. Chinese tasks are longer on average than English ones (41.13 vs.\ 34.08 steps), further reflecting the complexity of the targeted real-world setting.

\subsection{Benchmark Construction and Curation}
\label{sec:browserbench_construction}
BrowserBench is built from authentic browser-use scenarios rather than synthetic templates. Candidate tasks are collected from real Chinese and English websites and retained only when they require meaningful multi-step interaction, such as cross-page navigation, comparison across multiple sources, structured information extraction, or task completion under changing browser context.

A key design principle of BrowserBench is that long-horizon tasks should reflect executable real-web interactions rather than abstract templates. We therefore curate tasks directly from realistic browsing scenarios, normalize them into goal-oriented instructions paired with structured success criteria, and verify them through manual review and repeated browser execution before release.

To improve evaluation realism and interpretability, benchmark construction follows a strict curation process. We remove cases with excessive procedural leakage, ambiguous goals, unsupported entities, or stale page structures, so that benchmark failures are more likely to reflect model limitations rather than annotation noise or environment mismatch. All retained tasks are manually verified for executability, textual quality, and alignment between the task goal and the actual browser environment.

This curation process makes benchmark failures more informative: when an agent fails on BrowserBench, the failure is more likely to reflect weakness in planning, grounding, long-horizon interaction, or recovery, rather than noise in the benchmark itself.

\subsection{Task Taxonomy and Dataset Statistics}
\label{sec:browserbench_taxonomy}

BrowserBench covers eight task categories that reflect common patterns in real-world browser use:

\begin{itemize}
    \item \textbf{E-commerce and product information}: 164 tasks (46.9\%), including 98 Chinese and 66 English tasks.
    \item \textbf{Rankings and comparison}: 52 tasks (14.9\%), including 34 Chinese and 18 English tasks.
    \item \textbf{Data collection and batch extraction}: 52 tasks (14.9\%), including 16 Chinese and 36 English tasks.
    \item \textbf{Maps, travel, and tourism}: 30 tasks (8.6\%), including 18 Chinese and 12 English tasks.
    \item \textbf{Organization verification}: 21 tasks (6.0\%), including 13 Chinese and 8 English tasks.
    \item \textbf{Academic and knowledge search}: 16 tasks (4.6\%), including 6 Chinese and 10 English tasks.
    \item \textbf{General search and multi-step browsing}: 11 tasks (3.1\%), including 4 Chinese and 7 English tasks.
    \item \textbf{News and content browsing}: 4 tasks (1.1\%), including 2 Chinese and 2 English tasks.
\end{itemize}

Each task is additionally labeled with a difficulty level. Rather than
relying on subjective annotation, difficulty is calibrated by empirical
solvability: each task is attempted by three reference agents of distinct
capability tiers, none from the Qwen3.5 family used
by our own models. Tasks solved by at least two calibrators are labeled
\texttt{easy}, tasks solved by exactly one are \texttt{medium}, and tasks
solved by none are \texttt{hard}, followed by human review of boundary
cases. The resulting distribution is 105 easy (30\%), 140 medium (40\%),
and 105 hard (30\%) tasks; the hard share exceeds that of
Online-Mind2Web (26.3\%), giving BrowserBench greater discriminative
power at the difficulty ceiling. We further verify that the difficulty
split is balanced across the two languages and that average completion
length increases monotonically from easy to hard, so difficulty is not a
proxy for either language or length alone.

This taxonomy enables analysis beyond a single aggregate benchmark score. It provides interpretable slices over different real-web task structures and language settings, allowing us to examine whether a browser agent generalizes across not only websites, but also task patterns that differ substantially in planning, interaction, and extraction demands.

In addition to language and category coverage, BrowserBench is distinguished by its horizon distribution. The mean task length is 37.9 steps across the full benchmark, substantially longer than conventional short-browser benchmarks. Chinese tasks are on average more demanding than English tasks, and the benchmark includes trajectories up to 100 steps, making it suitable for studying precisely the long-horizon regime in which deviations, detours, and recovery become routine.

\subsection{Evaluation Protocol and Metric}
\label{sec:browserbench_protocol}

All BrowserBench evaluations follow a strict goal-only protocol: the agent receives only the normalized task instruction, with no auxiliary procedural guidance or hidden intermediate hints. A run is counted as successful only if the agent satisfies the corresponding structured success criterion within a single rollout. All runs use a maximum budget of 100 interaction steps; tasks not completed within the budget are counted as failures, reflecting the bounded interaction cost of realistic deployment.

We adopt Pass@1 Success Rate (SR) as the benchmark metric. Because each task is evaluated with exactly one rollout, Pass@1 is equivalent to single-rollout success rate. Let $\mathcal{C}$ denote the full set of benchmark cases. The overall benchmark score is defined as
\begin{equation}
    \mathrm{SR}_{\mathrm{total}}
    =
    \frac{1}{|\mathcal{C}|}
    \sum_{c \in \mathcal{C}} \mathbf{1}\{\mathrm{success}(c)\},
\end{equation}
where $\mathbf{1}\{\mathrm{success}(c)\}$ equals 1 if the agent successfully completes case $c$ and 0 otherwise.

We use Pass@1 Success Rate because realistic browser deployment requires reliable execution within a single trajectory rather than repeated sampling until success. This makes BrowserBench particularly suitable for measuring the robustness of long-horizon browser agents under realistic interaction constraints.

Overall, BrowserBench complements existing benchmarks by explicitly targeting the bilingual, long-horizon, real-web regime that remains underrepresented in current browser-agent research.

\section{A Self-Reinforcing Data Flywheel}\label{sec:flywheel}

As illustrated in Figure~\ref{fig:network_overview}, the training pipeline is embedded in a broader self-reinforcing data flywheel. The components introduced above are not merely a one-shot training recipe; together with the harness and BrowserBench, they form a closed data loop in which each round of deployment produces the supervision for the next. This property matters in the browser domain specifically: live websites change continuously, so any fixed demonstration corpus depreciates over time, and sustained capability requires a pipeline that converts ongoing interaction, including failed interaction, into fresh, verified training signal.

The flywheel operates in four stages. \textbf{(1) Unified trajectory collection.} Because SFT data construction, online rollouts, and evaluation all run on the same harness (Section~\ref{sec:harness}), every executed episode---whether from DAO-GRPO training, benchmark evaluation, or deployment---is logged in an identical, training-ready format with structured actions, feedback, and reconstructed contexts. \textbf{(2) Automated triage.} A step-level judge first separates genuine model failures from environment-induced ones such as transient pages or unreachable sites, routing the latter to task-pool maintenance rather than training; it then classifies the remainder: successful episodes become candidate general demonstrations for $D_g$, failed episodes with a localizable erroneous step become reflection candidates, and failures concentrated on specific widgets are routed to the UI-component pipeline as new $D_u$ scenes. In this way the loop recovers training signal from failed trajectories rather than discarding them. \textbf{(3) Verification-gated augmentation.} Candidate corrections are re-executed in the sandbox, and only strategies that verifiably restore the task to a viable path enter $D_r$ (Section~\ref{sec:reflection_data}); this gate prevents the loop from amplifying its own mistakes, a failure mode commonly observed in unverified self-training. \textbf{(4) Diagnostic targeting.} Error-category statistics from triage and tag-wise BrowserBench results are aggregated into a prioritized set of capability targets, which steer both task synthesis for the online pool $\mathcal{D}$ and data collection for the next SFT round, closing the loop from evaluation back to supervision.

The loop is automated where verification is cheap and human where judgment is scarce: trajectory logging, triage, and correction re-execution run without supervision, while humans author reflection reasoning, annotate complex-UI interactions, and audit borderline judge decisions. Evaluation signals are likewise split by their cost--precision profile: deterministic structured criteria where reproducibility matters (BrowserBench scoring), and a routed LLM judge where throughput matters (reward estimation over thousands of online rollouts).

In this work we instantiate the bootstrapping cycle of this flywheel: an initial supervised model trained on general demonstrations only (Base SFT in Table~\ref{tab:ablation_sft}) bootstraps roughly 3{,}000 rollout trajectories, from which about 420 verified reflection examples and 130 UI-component scenes are distilled and folded into the training mixture of the released models; diagnostic slices from an early BrowserBench snapshot guided the composition of the online task pool.

Because triage and correction re-execution are automated, each additional cycle requires only sandbox compute and lightweight human auditing rather than proportional annotation effort. We view sustained multi-cycle operation of this flywheel, including preference-based training over the accumulated contrastive pairs, as the primary mechanism for keeping Wuying-Browser-Agent aligned with the evolving web.

\section{Experiments}

\subsection{Training Settings}
\label{sec:training_settings}

\textbf{SFT.} We initialize the browser policy with RUIC-SFT. Starting from Qwen3.5 series models, we fine-tune for two epochs using LoRA~\cite{hu2022lora} with rank $16$, scaling factor $\alpha=64$, and all linear layers as target modules. Optimization uses a cosine learning-rate schedule with a peak learning rate of $5\times10^{-5}$ and a $10\%$ linear warmup. Training is conducted on a multi-GPU cluster with sequence parallelism enabled to accommodate long browser trajectories.

\textbf{Online RL.} We further optimize the policy with DAO-GRPO using parameter-efficient LoRA adaptation (rank $8$, $\alpha=16$, target modules: all linear layers). Online training runs against a large-scale AgentBay sandbox pool orchestrated by an asynchronous rollout scheduler: tasks are dispatched in batches, rolled out in parallel across isolated sandboxes, and streamed back to the optimizer upon completion. Sandboxes are recycled after each episode to guarantee state isolation. The reward combines format-validity checks with an LLM-as-a-judge success signal; a routed evaluation strategy keeps judging cost tractable at scale. Trajectories that terminate abnormally or become unreachable are masked out by setting their advantages to zero. Each rollout trajectory is capped at $50$ interaction steps as a throughput--coverage trade-off for online training rather than a deployment horizon limit: tasks exceeding this cap are truncated with the last observed state treated as non-terminal, retaining accumulated shaping progress as a learning signal. Evaluation always uses the full 100-step budget (Section~\ref{sec:browserbench_protocol}). Training proceeds with trajectory-level dynamic sampling on the held-out validation set, discarding degenerate groups with identical returns at each update.

\subsection{Evaluation Benchmarks}
\label{sec:benchmarks}

We evaluate Wuying-Browser-Agent on two groups of benchmarks covering browser-use capability and general agentic capability. For fair comparison, all evaluated models interact with web pages through the same harness, tool space, and observation pipeline; performance differences therefore reflect policy capability rather than differences in execution infrastructure.

\textbf{Browser-use benchmarks.} We use three online browser benchmarks. WebVoyager~\cite{he2024webvoyager} contains 643 tasks spanning 15 popular websites and is widely used for evaluating real-web navigation. Online-Mind2Web~\cite{xue2025illusion} contains 300 tasks across 136 websites and places greater emphasis on longer multi-step interaction under live web conditions. Our proposed BrowserBench contains 350 bilingual real-web tasks spanning 254 websites, with an average completion length of 37.9 steps, and is designed to evaluate browser agents in the Chinese--English long-horizon setting. Unless otherwise specified, all browser-use results are reported as Pass@1 Success Rate under the official evaluation protocol. The evaluation framework is built on a customized version of \texttt{browser-use}\footnote{\url{https://github.com/browser-use/browser-use}}. Since the same judge family is also used for reward estimation during
online training, we audited judge reliability on a random sample of 500 trajectories: judge decisions agree with independent human annotation in 96.4\% of cases, with no systematic bias favoring
Qwen-family policies.

\textbf{General agentic benchmarks.} To assess whether browser-grounded training transfers beyond web interaction, we additionally report results on Tau2-Bench~\cite{barres2506tau2}, BFCL-V4~\cite{patil2025berkeley}, and Claw-Eval~\cite{ye2026claw}. Tau2-Bench evaluates multi-domain conversational tool use, BFCL-V4 focuses on function-calling reliability in both single-turn and multi-turn settings, and Claw-Eval measures end-to-end autonomous agent performance on human-verified tasks. Together, they provide a complementary view of whether improvements for long-horizon browser interaction are achieved without sacrificing broader agentic capability.

\subsection{Compared Baselines}
\label{sec:baselines}

We compare Wuying-Browser-Agent against the baselines listed in Table~\ref{tab:official_success_rates} and Table~\ref{tab:agentic_selected_transposed}.

\textbf{Browser-use baselines.} For browser-use benchmarks, we divide baselines into closed-source and open-source models following Table~\ref{tab:official_success_rates}. The closed-source baselines are GPT-4o~\cite{hurst2024gpt}, GPT-5~\cite{singh2025openai}, GPT-5.5~\cite{openai2026gpt55}, Ernie-5.0~\cite{wang2026ernie}, and Seed2.1-Pro~\cite{bytedance2026seed2}. The open-source baselines are OpenWebRL-8B~\cite{yang2026openwebrl}, Kimi k2.5~\cite{team2026kimi}, Qwen3-VL-8B-Thinking~\cite{bai2025qwen3}, Qwen3-VL-235B-A22B-Thinking~\cite{bai2025qwen3}, Hy3-295B-A21B~\cite{tencent2025hy3}, Qwen3.5-4B/9B/27B/397B-A17B~\cite{yang2025qwen3}, Qwen3.8-Max, and DeepSeek-V4-Flash-0731~\cite{xu2026deepseek}. These baselines cover browser-specialized systems as well as general-purpose multimodal foundation models across a wide range of scales.

\textbf{General agentic baselines.} For general agentic benchmarks, we compare against the models reported in Table~\ref{tab:agentic_selected_transposed}. Browser-agent baselines include OpenCUA 72B~\cite{wang2026opencua}, GUI-Owl-1.5-32B-Instruct~\cite{xu2026mobile}, UI-Venus-1.5-30B-A3B~\cite{team2026ui}, and EvoCUA-32B-20260105~\cite{huang2026evocua}. General-purpose agent baselines include Qwen3.5-27B~\cite{yang2025qwen3} and Qwen-UI-Agent-27B~\cite{zhou2026qwen}. This comparison allows us to assess whether the browser-grounded training of Wuying-Browser-Agent transfers beyond browser interaction to broader tool-use and autonomous-agent settings.


\definecolor{avgblue}{RGB}{220,230,241}
\definecolor{lightpurple}{rgb}{0.8, 0.6, 1.0}

\begin{table}[h]
\centering
\caption{Task success rates (\%) on three open-web benchmarks: WebVoyager, Online-Mind2Web, and BrowserBench. All models are evaluated with a maximum budget of 100 interaction steps. The Average column is the arithmetic mean of the three benchmarks. The
best result in each column is shown in bold, and the second-best is
underlined.}
\label{tab:official_success_rates}
\setlength{\tabcolsep}{6pt}
\renewcommand{\arraystretch}{1.15}
\resizebox{0.86\columnwidth}{!}{
\begin{tabular}{l|ccc|>{\columncolor{lightpurple!30}}c}
\toprule
\textbf{Model Name} & \textbf{WebVoyager} & \textbf{Online-Mind2Web} & \textbf{BrowserBench} & \textbf{Average} \\
\midrule

\multicolumn{5}{l}{\textit{Closed-Source Models}} \\
\midrule
GPT-4o & 48.7 & 27.9 & 22.0 &32.9 \\
GPT-5 & 80.1 & 67.1 & 61.7 &69.6 \\
GPT-5.5 & \textbf{85.1} & \textbf{74.7} & \textbf{67.4} & \textbf{75.7}\\
Ernie-5.0 (>1T) & 51.9 & 31.3 & 24.9 & 36.0\\
Qwen3.7-Plus & 73.6 & 60.1 & 59.4 & 64.4\\
Seed2.1-Pro & 70.6 & 59.2 &40.2  &56.7 \\
\midrule
\multicolumn{5}{l}{\textit{Open-Source Models}} \\
\midrule
OpenWebRL-8B & 56.7 & 45.6 & 34.3 &45.5 \\
Kimi2.5 & 71.7 & 57.1 & 59.7 & 62.8\\
Qwen3-VL-235B-A22B-Thinking & 57.2 & 44.9 &  30.9&44.3 \\
Hy3-295B-A21B & 61.4 & 44.1 & 30.3 & 45.3\\
Qwen3.5-4B & 37.6 & 20.0 & 16.3 &24.6 \\
Qwen3.5-9B & 37.0 & 23.0 & 22.9 & 27.6\\
Qwen3.5-27B & 55.4 & 50.0 & 37.1 &47.5 \\
Qwen3.5-397B-A17B & 70.8 & 54.4 &44.3  &56.5 \\
Qwen3.8-Max (2.4T-A95B) & 77.8 & \underline{68.9} & 64.3 &70.3  \\
DeepSeek-V4-Flash-0731 (284B-A13B)& 69.0 & 56.8 & 48.9 &58.2  \\
\midrule

\multicolumn{5}{l}{\textit{Ours: 4B backbone}} \\
\midrule
\textbf{Wuying-Browser-Agent-4B-SFT} & 50.0 & 35.1 & 24.3 & 36.5\\
\textbf{Wuying-Browser-Agent-4B} & 56.5 & 41.7 &  34.0& 44.1\\
\midrule

\multicolumn{5}{l}{\textit{Ours: 9B backbone}} \\
\midrule
\textbf{Wuying-Browser-Agent-9B-SFT} & 58.0 & 40.7 & 38.0 &45.6 \\
\textbf{Wuying-Browser-Agent-9B} & 64.0 & 45.5 &42.9  &50.8 \\
\midrule

\multicolumn{5}{l}{\textit{Ours: 27B backbone}} \\
\midrule
\textbf{Wuying-Browser-Agent-27B-SFT} & 73.8 & 59.5 & 54.9 &62.7 \\
\textbf{Wuying-Browser-Agent-27B} & \underline{80.6} & 66.7 & \underline{65.1} &\underline{70.8} \\
\bottomrule
\end{tabular}}
\end{table}

\subsection{Browser-Use Benchmark Results}
\label{sec:main_results}

Table~\ref{tab:official_success_rates} summarizes task success rates on WebVoyager, Online-Mind2Web, and BrowserBench. Overall, Wuying-Browser-Agent establishes a new open-source state of the art on browser-use benchmarks and remains competitive with strong closed-source systems.

Among open-source models, Wuying-Browser-Agent-27B achieves the best average success rate at 70.8\%, slightly exceeding the strongest competing open model, Qwen3.8-Max (70.3\%), and substantially outperforming Qwen3.5-397B-A17B (56.5\%). This advantage is consistent across the three browser-use benchmarks, where Wuying-Browser-Agent-27B reaches 80.6\% on WebVoyager, 66.7\% on Online-Mind2Web, and 65.1\% on BrowserBench. Taken together, these results suggest that the proposed pipeline improves browser-agent performance across heterogeneous live-web settings rather than overfitting to a single benchmark style.

The three benchmarks are also complementary in coverage. WebVoyager and Online-Mind2Web are widely used English-centric live-web benchmarks, whereas BrowserBench places greater emphasis on Chinese-web and more deployment-oriented real-world websites. Strong performance across all three therefore indicates that Wuying-Browser-Agent generalizes across both benchmark conventions and web environments with different linguistic and interaction characteristics.

Wuying-Browser-Agent-27B is also competitive with strong closed-source systems. Its average score surpasses Qwen3.7-Plus (70.8\% vs.\ 64.4\%) and remains reasonably close to GPT-5 (69.6\%) and GPT-5.5 (75.7\%). This result is encouraging given that Wuying-Browser-Agent is fully open-source and trained with a unified pipeline explicitly targeted at long-horizon real-browser interaction.

We further observe that the gains from online RL are consistent across model scales. At 9B, the final model improves over its SFT counterpart from 45.6\% to 50.8\%; at 27B, performance rises from 62.7\% to 70.8\%. This pattern suggests that DAO-GRPO provides stable gains beyond supervised initialization rather than benefiting only a specific scale. Overall, both the SFT and RL variants improve with model size, indicating that the proposed RUIC-SFT and DAO-GRPO pipeline can effectively convert increased model capacity into stronger browser-agent capability.

\subsection{Transfer to General Agentic Benchmarks}
\label{sec:agentic_transfer}
Table~\ref{tab:agentic_selected_transposed} evaluates whether the browser-grounded training of Wuying-Browser-Agent transfers beyond browser use to more general agentic settings. Across Tau2-Bench, Claw-Eval, and BFCL-v4, Wuying-Browser-Agent-27B remains competitive with strong general-purpose tool agents rather than over-specializing to web interaction. In particular, it improves over the Qwen3.5-27B base model on all reported benchmarks, indicating that the gains obtained from harness-grounded supervision and long-horizon browser optimization do not come at the expense of broader agentic capability. Compared with Qwen-UI-Agent-27B, Wuying-Browser-Agent-27B is also competitive overall while being trained primarily for real-web browser interaction. These results suggest that the proposed pipeline strengthens not only browser-specific robustness, but also more general abilities in structured tool use, multi-turn task execution, and autonomous action under feedback.

\begin{table}[t]
\centering
\caption{Agentic performance comparison on selected benchmarks. We report performance on Tau2-Bench, Claw-Eval, and BFCL-v4 to evaluate whether browser-grounded training preserves broader tool-use and autonomous-agent capability beyond browser-specific tasks.}
\label{tab:agentic_selected_transposed}
\resizebox{\linewidth}{!}{
\begin{tabular}{l|cccc|c}
\toprule
\textbf{Model} & \textbf{Tau2-Bench} & \textbf{Claw-Eval (Avg 3)} & \textbf{Claw-Eval (Pass$^3$)} & \textbf{BFCL-v4} & \textbf{Average} \\
\midrule
\multicolumn{6}{l}{\textit{Browser Agents}} \\
\midrule
OpenCUA 72B                 & 14.4 & 26.4 & 0.5  & 28.3 & 17.4 \\
GUI-Owl-1.5-32B-Instruct    & 6.1  & 29.6 & 5.5  & 32.7 & 18.5 \\
UI-Venus-1.5-30B-A3B        & 22.7 & 30.6 & 5.5  & 19.8 & 19.7 \\
EvoCUA-32B-20260105         & 48.9 & 46.3 & 6.5  & 48.8 & 37.6 \\
\midrule
\multicolumn{6}{l}{\textit{General Models}} \\
\midrule
Qwen3.5-27B                 & 89.2 & 66.9 & 41.2 & 71.3 & 67.2 \\
Qwen-UI-Agent-27B           & 89.9 & 73.5 & 51.8 & 74.2 & 72.4 \\
Wuying-Browser-Agent-27B & \textbf{91.6} &\textbf{74.1}  & \textbf{52.8} & \textbf{76.5} & \textbf{73.8} \\
\bottomrule
\end{tabular}
}
\end{table}

\subsection{Ablation Study}
\label{sec:ablation}

We conduct ablation studies to isolate the contribution of each component in RUIC-SFT and DAO-GRPO. All ablations use the 9B backbone
and are evaluated on BrowserBench unless otherwise stated.

\subsubsection{Ablation on RUIC-SFT}
\label{sec:ablation_sft}

Table~\ref{tab:ablation_sft} studies the contribution of the two specialized data sources and the curriculum schedule in RUIC-SFT. The results show that $D_u$ and $D_r$ provide complementary benefits. Compared with Base SFT (\#1), adding $D_u$ (\#2) improves the overall success rate from 32.0\% to 34.9\% while keeping the recovery success rate nearly unchanged (8.5\% vs.\ 9.0\%), indicating that UI-specialized supervision primarily strengthens browser-operation competence rather than error correction. In contrast, adding $D_r$ (\#3) nearly doubles the recovery success rate from 8.5\% to 16.4\% together with a comparable overall gain, showing that reflection data is particularly effective for off-trajectory correction and self-recovery.

Combining all three data sources with a fixed global mixture (\#4) further improves overall success rate over the single-source variants, but also increases the average number of actions per completed task from 24.8 to 29.3. This suggests that naive uniform mixing may introduce more redundant or hesitant behavior during execution, despite improving task completion.

In contrast, the full RUIC-SFT curriculum (\#5) achieves the best overall performance, improving success rate to 38.0\% and recovery success rate to 18.5\%, while reducing the average step count to 22.4. This indicates that phased curriculum scheduling better balances execution efficiency and recovery behavior than fixed-ratio multi-source training.

Finally, introducing reflection data aggressively from the beginning of training (\#6) degrades overall performance relative to both the fixed mixture and the full curriculum, despite maintaining a recovery success rate comparable to the fixed-mixing variant. This supports our design choice of delaying reflection-heavy supervision until a stable browser-operation prior has first been established.

\begin{table}[t]
\centering
\caption{\textbf{Ablation on RUIC-SFT.} All variants use the same 9B backbone and are evaluated on BrowserBench. ``Fixed'' applies a fixed multi-source mixture uniformly throughout training, while ``Curriculum'' uses the phased schedule. SR: overall task success rate; Recov. SR: erroneous-step recovery rate;  Steps: average number of actions per completed task.}
\label{tab:ablation_sft}
\setlength{\tabcolsep}{5pt}
\renewcommand{\arraystretch}{1.15}
\small
\begin{tabular}{@{}c l cccc ccc@{}}
\toprule
\multirow{2}{*}{\textbf{\#}} & \multirow{2}{*}{\textbf{Variant}}
  & \multicolumn{3}{c}{\textbf{Data Sources}}
  & \multirow{2}{*}{\textbf{Schedule}}
  & \multirow{2}{*}{\textbf{SR (\%)}}
  & \multirow{2}{*}{\textbf{Recov. SR (\%)}}
  & \multirow{2}{*}{\textbf{Steps}} \\
\cmidrule(lr){3-5}
  & & $D_g$ & $D_u$ & $D_r$ & & & & \\
\midrule
1 & Base SFT
  & $\cmark$ & $\xmark$ & $\xmark$ & None
  & 32.0 & 8.5  & 24.8 \\
2 & +UI
  & $\cmark$ & $\cmark$ & $\xmark$ & Fixed
  & 34.9 & 9.0 &  23.1 \\
3 & +Reflection
  & $\cmark$ & $\xmark$ & $\cmark$ & Fixed
  & 35.5 & 16.4 &  27.6 \\
4 & +UI+Reflection
  & $\cmark$ & $\cmark$ & $\cmark$ & Fixed
  & 36.6 & 17.1 &  29.3 \\
5 & \textbf{RUIC-SFT (full)}
  & $\cmark$ & $\cmark$ & $\cmark$ & Curriculum
  & \textbf{38.0} & \textbf{18.5} & 22.4 \\
6 & Early-Reflection
  & $\cmark$ & $\cmark$ & $\cmark$ & Aggressive Early Ref.$^\dagger$
  & 35.1 & 16.6 & 30.5  \\
\bottomrule
\end{tabular}
\vspace{-2pt}
\begin{minipage}{\linewidth}
\footnotesize
$^\dagger$ Reflection data is introduced at a high initial ratio from the beginning of training and later reduced significantly, testing the effect of excessive early reflection on policy stabilization.
\end{minipage}
\end{table}

\subsubsection{Ablation on DAO-GRPO}
\label{sec:ablation_daograpo}

Table~\ref{tab:ablation_daograpo} evaluates the contribution of the main design choices in our online RL framework, including denser progress supervision, more targeted credit assignment, and training under step-wise decision contexts. All variants are initialized from the same RUIC-SFT checkpoint and trained for the same number of online iterations.

Vanilla online GRPO improves only modestly over the RUIC-SFT initialization (38.0\% $\rightarrow$ 39.4\%), confirming that sparse terminal rewards alone provide limited supervision for long-horizon browser tasks. Adding PBRS improves the overall success rate to 40.7\%, and also yields gains on hard tasks and recovery-oriented tasks. This result suggests that dense progress signals are beneficial in browser environments where terminal rewards are sparse and meaningful intermediate progress is otherwise weakly supervised.

Adding divergence-aware credit assignment further improves performance, with the largest gains appearing exactly where the method targets: the hard subset (25.7\% $\rightarrow$ 27.6\%) and recovery-oriented tasks (26.2\% $\rightarrow$ 29.7\%). This supports our hypothesis that browser trajectories often share long prefixes and only diverge at a few decisive branch points, making localized credit assignment more effective than uniformly weighting all response segments.

Finally, enabling the full response-level objective under reconstructed contexts further improves all metrics, raising the overall success rate to 42.9\%, the hard-task success rate to 30.5\%, and the recovery-oriented success rate to 32.8\%. This result confirms the importance of optimizing each response under its own decision context, rather than under a single concatenated trajectory transcript, since browser-agent interaction is naturally state-reconstructed rather than append-only.

Overall, the ablation validates that the three components are complementary: PBRS alleviates sparse supervision, divergence-aware credit sharpens optimization around decisive branch steps, and response-level optimization aligns the learning objective more faithfully with actual browser inference.

\begin{table}[t]
\centering
\caption{\textbf{Ablation on DAO-GRPO components.} All variants are initialized from the same RUIC-SFT checkpoint and trained for the same number of online iterations. SR: overall task success rate; Hard SR: success rate on hard BrowserBench tasks; Recov. SR: success rate on recovery-oriented tasks.}
\label{tab:ablation_daograpo}
\setlength{\tabcolsep}{8pt}
\renewcommand{\arraystretch}{1.15}
\resizebox{0.99\columnwidth}{!}{
\begin{tabular}{l ccc ccc}
\toprule
\multirow{2}{*}{\textbf{Variant}}
  & \multirow{2}{*}{\textbf{PBRS}}
  & \multirow{2}{*}{\textbf{Div. Credit}}
  & \multirow{2}{*}{\textbf{Resp.-Level}}
  & \multirow{2}{*}{\textbf{SR (\%)}}
  & \multirow{2}{*}{\textbf{Hard SR (\%)}}
  & \multirow{2}{*}{\textbf{Recov. SR (\%)}} \\
  & & & & & & \\
\midrule
Vanilla Online GRPO
  & $\xmark$ & $\xmark$ & $\xmark$
  & 39.4 & 23.8 & 24.8 \\
+PBRS
  & $\cmark$ & $\xmark$ & $\xmark$
  & 40.7 & 25.7 & 26.2 \\
+PBRS+Div. Credit
  & $\cmark$ & $\cmark$ & $\xmark$
  & 41.9 & 27.6 & 29.7 \\
\textbf{DAO-GRPO (full)}
  & $\cmark$ & $\cmark$ & $\cmark$
  & \textbf{42.9} & \textbf{30.5} & \textbf{32.8} \\
\bottomrule
\end{tabular}}
\end{table}

\subsubsection{Reliability of the Divergence Estimator}
\label{sec:estimator_reliability}

\begin{figure}[t]
    \centering
    \includegraphics[width=0.98\textwidth]{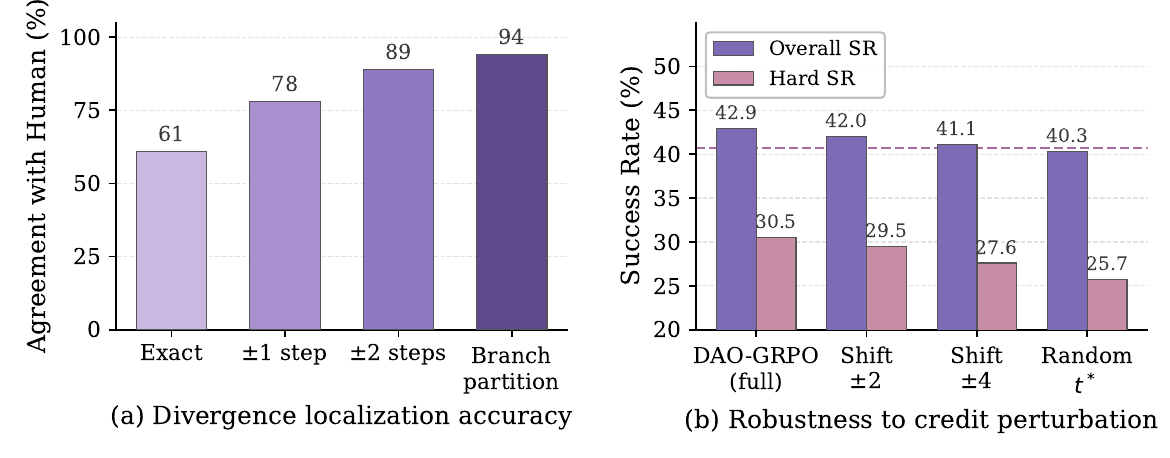}
        \caption{Reliability of the LLM-based divergence estimator.
    (a) Agreement between estimator outputs and human annotation on 100
    labeled rollout groups: exact / $\pm$1-step / $\pm$2-step
    localization of $t^*$ and favorable--unfavorable branch partition.
    (b) BrowserBench success rate of DAO-GRPO (9B) when the detected
    $t^*$ is randomly perturbed by up to $\pm$2 or $\pm$4 steps or
    replaced by a random step; the dashed line marks the uniform-credit
    variant (+PBRS, 40.7\%) from Table~\ref{tab:ablation_daograpo}.}
    \label{fig:estimator_reliability}
\end{figure}

Since divergence-aware credit assignment relies on an LLM-based estimator,
we validate its reliability directly. Two annotators independently labeled
the ground-truth divergence step $t^*$ on 100 rollout groups sampled from
training (inter-annotator agreement within one step: 91\%). As shown in
Figure~\ref{fig:estimator_reliability}(a), the estimator localizes $t^*$
exactly in 61\% of groups, within one step in 78\%, and within two steps
in 89\%, while the favorable/unfavorable branch partition agrees with
human annotation in 94\% of groups. The estimator is thus approximately
correct rather than exact, which matches its intended role as a credit
prior (Section~\ref{sec:credit}).

We further test whether DAO-GRPO tolerates this level of localization
noise. We retrain the policy under the identical schedule while
perturbing every detected $t^*$ by a random offset of up to $\pm2$ or
$\pm4$ steps, or replacing it with a uniformly random step. As shown in
Figure~\ref{fig:estimator_reliability}(b), performance degrades
gracefully: $\pm2$ perturbation costs $0.9$ points (42.9\%
$\rightarrow$ 42.0\%) and even $\pm4$ perturbation retains 41.1\%,
still above the uniform-credit variant (+PBRS, 40.7\%), indicating that
the method only requires the divergence prior to be approximately
correct. Random placement drops below uniform credit (40.3\%),
confirming that concentrating updates at wrong steps is worse than not
concentrating at all; degradation is amplified on the hard subset
(30.5\% $\rightarrow$ 25.7\%), where accurate branch localization
matters most.

\subsubsection{Reliability of the Progress Estimator $\Phi$}
\label{sec:phi_reliability}

The remaining LLM-based component is the progress estimator $\Phi$, which
converts trajectory prefixes into the potential values used by PBRS. We
validate it against human judgment on 150 trajectory prefixes sampled from
held-out dev rollouts, each rated by two annotators on a five-level
progress rubric anchored by subgoal descriptions (inter-annotator
agreement: 93\% within one level, Spearman $\rho = 0.89$). $\Phi$ tracks
the human ratings closely (Spearman $\rho = 0.81$, mean absolute error
$0.11$ on the unit interval), agrees with the human progress ordering on
86\% of 200 same-task prefix pairs, and is monotone non-decreasing along
87\% of adjacent within-trajectory steps. Since PBRS only requires the
potential to rank prefixes by progress rather than to be exact, this
level of agreement suffices for the shaping signal to be informative,
consistent with the net gain of the +PBRS variant in
Table~\ref{tab:ablation_daograpo}. Moreover, potential-based shaping
preserves the optimal policy for any bounded $\Phi$, so residual
estimator noise degrades credit quality gracefully rather than biasing
the optimization objective.

\begin{figure}[t]
    \centering
    \begin{subfigure}[b]{0.24\textwidth}
        \centering
        \includegraphics[width=\textwidth]{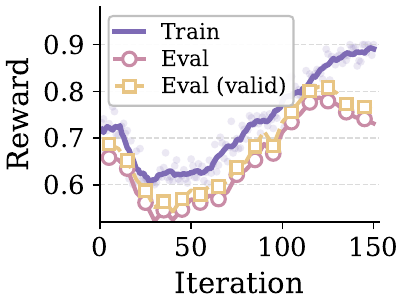}
        \caption{\textbf{Reward Trends}}
        \label{fig:online_rl_dynamics_reward}
    \end{subfigure}
    \hfill
    \begin{subfigure}[b]{0.24\textwidth}
        \centering
        \includegraphics[width=\textwidth]{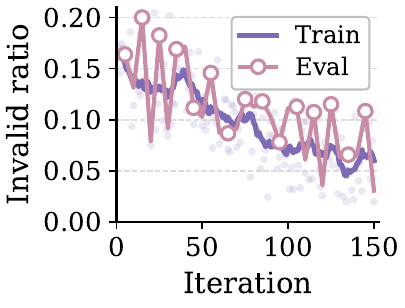}
        \caption{\textbf{Invalid Actions}}
        \label{fig:online_rl_dynamics_validity}
    \end{subfigure}
    \hfill
    \begin{subfigure}[b]{0.24\textwidth}
        \centering
        \includegraphics[width=\textwidth]{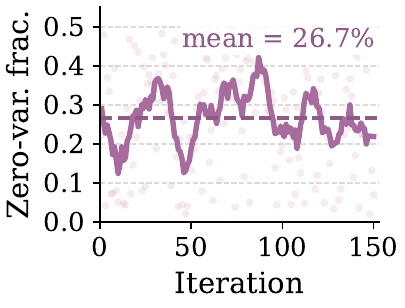}
        \caption{\textbf{Zero-Variance Groups}}
        \label{fig:online_rl_dynamics_zerovar}
    \end{subfigure}
    \hfill
    \begin{subfigure}[b]{0.24\textwidth}
        \centering
        \includegraphics[width=\textwidth]{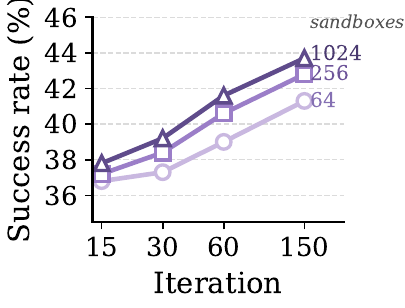}
        \caption{\textbf{Rollout Scaling}}
        \label{fig:online_rl_dynamics_scaling}
    \end{subfigure}

    \caption{
    Online RL training dynamics of Wuying-Browser-Agent-9B.
    (a) Shaped reward on training batches (per-step scatter with moving average) and on a held-out validation split, the latter also reported over valid rollouts only.
    (b) Invalid-action ratio on the training and validation environments throughout online RL.
    (c) Fraction of rollout groups with zero reward variance in each iteration; these groups contribute no relative-advantage signal and are masked by dynamic sampling. The dashed line marks the training average of 26.7\%.
    (d) Final success rate under different training budgets (up to 150 iterations) and parallel sandbox concurrency levels.
    Overall, online optimization improves both reward and action validity, degenerate groups persist throughout training and are handled by dynamic sampling, and higher sandbox concurrency yields better final performance under the same number of update steps.
    }
    \label{fig:online_rl_dynamics}
\end{figure}

\subsubsection{Online RL Training Dynamics}
Figure~\ref{fig:online_rl_dynamics} presents the online RL behavior of
DAO-GRPO on Wuying-Browser-Agent-9B. As shown in
Figure~\ref{fig:online_rl_dynamics_reward}, both the training reward and
the held-out validation reward exhibit an initial decline -- bottoming
out around iteration 30 -- followed by sustained improvement, with the
training reward recovering to its initial level by roughly iteration 90.
We attribute this early dip to the transition from the RUIC-SFT
initialization to online policy optimization, where the agent must first
adapt to exploration, grouped relative rewards, and dynamically
reconstructed browser contexts. After this adaptation phase, the reward
increases steadily, indicating that direct browser interaction provides
useful supervision beyond offline demonstrations. Importantly, the
validation reward computed on valid rollouts only follows the same
upward trend and remains consistently above the raw validation reward
(0.76 vs.\ 0.73 at iteration 150), showing that the gain is not merely
due to changes in invalid-rollout frequency.

Figure~\ref{fig:online_rl_dynamics_validity} further shows that the
invalid-action ratio consistently decreases on both the training and
validation environments (15.1\% $\rightarrow$ 6.1\% and 16.4\%
$\rightarrow$ 3.1\%, respectively, comparing the first and last ten
iterations). This suggests that DAO-GRPO improves not only task-level
return but also the executability and reliability of browser actions,
which is critical for real-world browser-agent deployment.

Figure~\ref{fig:online_rl_dynamics_zerovar} reports the fraction of
rollout groups with zero reward variance, i.e., groups in which all
rollouts in a group obtain identical returns and therefore provide no
relative-advantage signal. Such degenerate groups are frequent --
26.7\% of all groups on average and up to 49.9\% in individual
iterations -- and, notably, their frequency does not decrease as the
policy improves (21.2\% in the first ten iterations vs.\ 22.2\% in the
last ten), because they arise predominantly from tasks that remain
unsolvable rather than from early-stage format failures. Dynamic
sampling masks these groups from every update, so the effective batch
is substantially smaller than the nominal one throughout training;
this persistent filtering is what makes the rollout-throughput
considerations below practically important rather than incidental.

Figure~\ref{fig:online_rl_dynamics_scaling} studies the effect of
sandbox concurrency under the asynchronous rollout scheduler. Under all
training budgets, higher concurrency improves final performance, and
the benefit widens as training proceeds. Note that the group size is
fixed throughout training; concurrency instead affects optimization
through data freshness and selectivity. With more sandboxes active,
completed trajectory groups are returned faster relative to policy
updates, so each update consumes rollouts generated by a more recent
policy, and dynamic sampling can draw from a larger pool of candidate
groups when discarding degenerate ones. Together, these results
validate that online RL yields complementary gains over RUIC-SFT and
that rollout throughput is itself an important factor for effective
browser-agent optimization.

\begin{figure}
    \centering
    \begin{subfigure}[b]{0.48\textwidth}
        \centering
        \includegraphics[width=\textwidth]{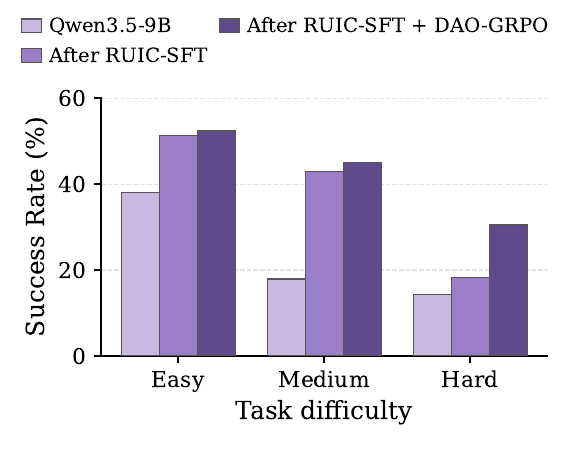}
        \caption{\textbf{Success Rate by Task Difficulty}}
        \label{fig:difficulty}
    \end{subfigure}
    \hfill
    \begin{subfigure}[b]{0.48\textwidth}
        \centering
        \includegraphics[width=\textwidth]{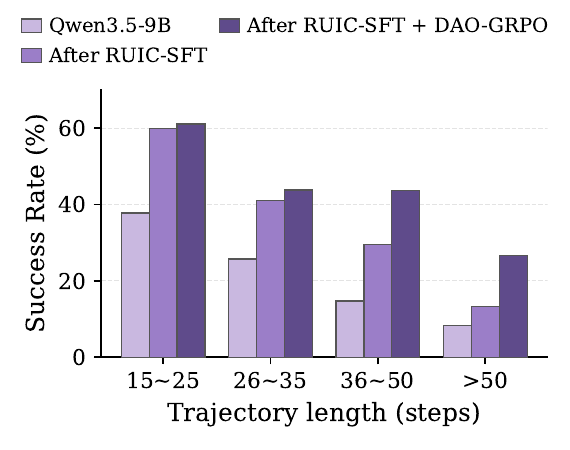}
        \caption{\textbf{Success Rate by Trajectory Length}}
        \label{fig:length}
    \end{subfigure}
    
    \caption{
    Success rate by task difficulty (a) and (b) trajectory
    length in steps on 9B model backbone.
    Trajectory length is binned into four ranges: 15--25, 26--35,
    36--50, and $>$50 steps.
    }
    \label{fig:difficulty_length}
\end{figure}

\subsubsection{Performance by Task Difficulty and Trajectory Length}
\label{sec:difficulty_analysis}

Figure~\ref{fig:difficulty_length} analyzes success rate by task difficulty (a) and trajectory length (b) for Qwen3.5-9B, RUIC-SFT, and DAO-GRPO.

In the left panel, performance decreases consistently from easy to medium to hard tasks for all three models, confirming that the empirical solvability-based difficulty labels align well with the actual challenge faced by the browser agent. RUIC-SFT improves over the base Qwen3.5-9B model across all difficulty levels. The largest gains appear on easy and medium tasks (38.1\% $\rightarrow$ 51.4\% and 17.9\% $\rightarrow$ 42.9\%), reflecting the browser-operation and UI-interaction competence instilled by curriculum supervision, while the gain on hard tasks is smaller (14.3\% $\rightarrow$ 18.1\%), as these tasks additionally demand long-horizon credit assignment that offline supervision alone cannot provide.

DAO-GRPO further improves performance at all difficulty levels, and its gains grow steeply with task difficulty: $+1.0$ point on easy tasks, where the supervised policy is already near saturation, $+2.1$ points on medium tasks, and $+12.4$ points on hard tasks (18.1\% $\rightarrow$ 30.5\%), the largest improvement across all slices. This suggests that online optimization is especially beneficial once the task requires long-horizon correction, branch-sensitive decision making, and recovery from off-trajectory states beyond what can be fully covered by offline supervision.

In the right panel, success rate drops as trajectory length increases, indicating that longer browser interactions are substantially more challenging. RUIC-SFT improves performance across all trajectory-length bins, with gains concentrated on shorter tasks (e.g., $+22.2$ points on 15--25 step tasks), showing that better browser-operation priors and recovery-oriented supervision are broadly useful but insufficient for the longest interactions. More importantly, the gain of DAO-GRPO becomes increasingly pronounced as trajectories grow longer: from $+1.1$ points on 15--25 step tasks to $+2.8$ points on 26--35 step tasks, $+5.2$ points on 36--50 step tasks, and $+13.4$ points on tasks exceeding 50 steps (13.3\% $\rightarrow$ 26.7\%). This pattern is well aligned with the motivation of DAO-GRPO: as browser trajectories become longer, sparse terminal reward, delayed credit assignment, and semantically decisive branch steps become increasingly important, making online RL particularly effective.

Overall, the figure shows that RUIC-SFT primarily strengthens difficult browser capabilities, while DAO-GRPO further improves long-horizon decision quality and recovery under challenging interaction settings.

\subsubsection{Case Study}

\begin{figure}[t]
    \centering
    \includegraphics[width=0.99\textwidth]{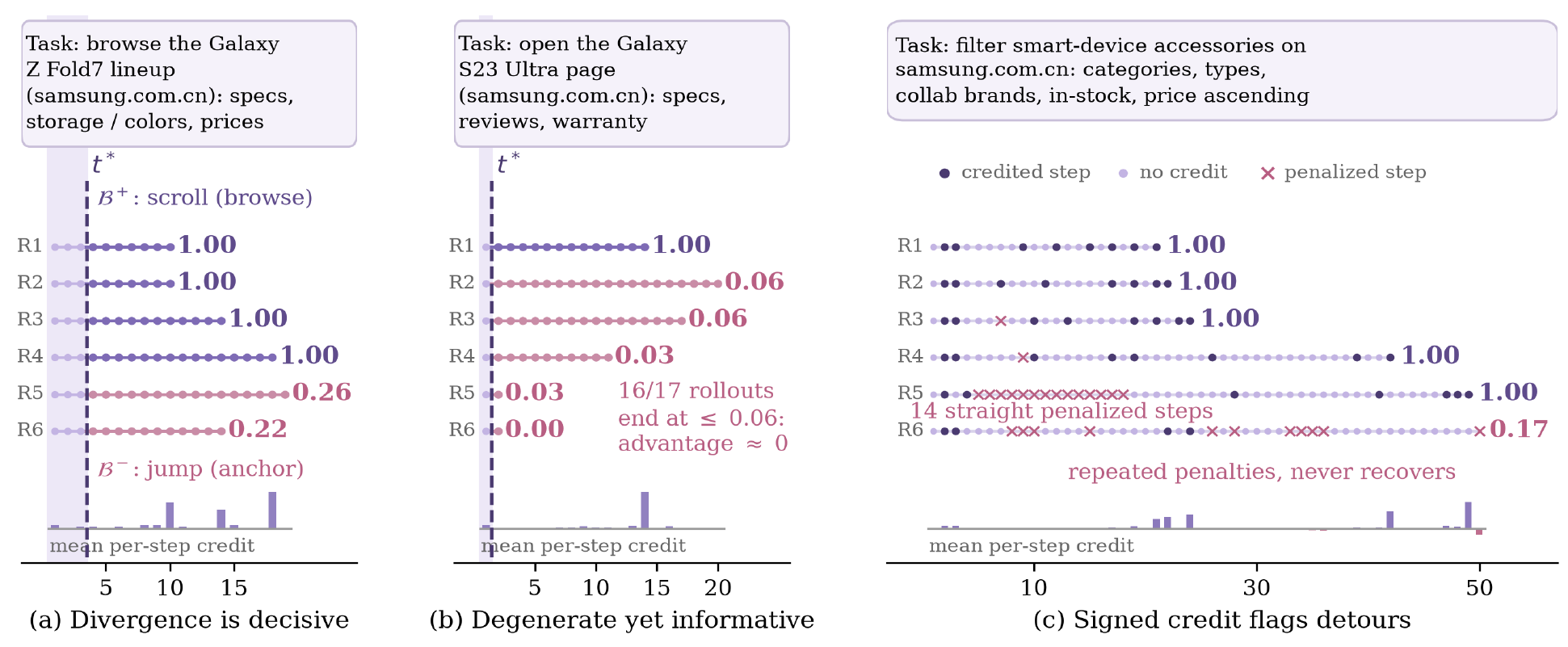}
    \caption{\textbf{Real rollout groups from online RL training,
    illustrating the three design choices of DAO-GRPO.} For analysis we
    sample 16--20 rollouts per task (beyond the training group size
    $K{=}6$) and display six representative ones per group; row ends show
    final rewards, and the bottom strips show the mean per-step PBRS
    shaped reward of the displayed rollouts.
    \textbf{(a)} A Galaxy Z Fold7 browsing group: all rollouts share an
    identical three-step prefix (shaded corridor) and diverge at
    $t^*{=}4$ into top-down browsing ($\mathcal{B}^+$: \texttt{scroll})
    vs.\ anchor jumping ($\mathcal{B}^-$: \texttt{scroll\_to\_text}); 12
    of 16 browsing rollouts succeed while 3 of 4 jumping rollouts stall
    at partial credit, and progress credit concentrates on the
    post-divergence steps of the favorable branch.
    \textbf{(b)} A Galaxy S23 Ultra group in which 16 of 17 rollouts fail
    with returns $\leq 0.06$: under terminal-only rewards the group would
    be degenerate (advantage $\approx 0$), whereas PBRS progress
    increments keep the rollouts correctly ordered ($1.00 > 0.06 > 0.03
    > 0.00$).
    \textbf{(c)} An accessory-filtering group in which 12 of 16 rollouts
    all reach reward $1.0$ with lengths from 21 to 49 steps; signed PBRS
    increments penalize individual detour steps (crosses; deep dots:
    credited steps; light dots: no credit), e.g.\ 14 consecutive
    penalized steps in R5, separating efficient from wasteful executions
    that identical terminal rewards cannot distinguish.}
    \label{fig:case_study_groups}
\end{figure}

We first examine three rollout groups sampled from online RL training
(Figure~\ref{fig:case_study_groups}), which illustrate why the three
components of DAO-GRPO match the structure of browser-agent data.
Figure~\ref{fig:case_study_groups}(a) exemplifies the shared-prefix
structure that motivates divergence-aware credit assignment:
trajectories within a group are identical for the first three steps and
diverge at a single decisive step, where the branch choice largely
determines the outcome. Top-down browsing succeeds in 12 of 16 cases,
whereas jumping directly to the product anchor stalls at partial credit
in 3 of 4 cases. The shaped credit therefore concentrates on the
post-divergence steps of the favorable branch.
Figure~\ref{fig:case_study_groups}(b) shows a group in which 16 of 17
rollouts fail with returns of at most $0.06$. Under terminal-only
rewards, such a group has zero advantage variance and would be
discarded by dynamic sampling
(Figure~\ref{fig:online_rl_dynamics_zerovar}), whereas PBRS progress
increments keep the rollouts correctly ordered and the group
informative.
Figure~\ref{fig:case_study_groups}(c) shows that even when all displayed
rollouts reach the same terminal reward, their efficiency differs by
more than a factor of two, ranging from 21 to 49 steps. Signed PBRS
increments penalize individual detour steps. For example, one eventually
successful rollout receives fourteen consecutive penalties during a
filter re-entry loop, providing step-level supervision toward efficient
behavior that identical terminal rewards cannot express.

\begin{figure}
    \centering \includegraphics[width=0.99\textwidth,height=1.0\textwidth]{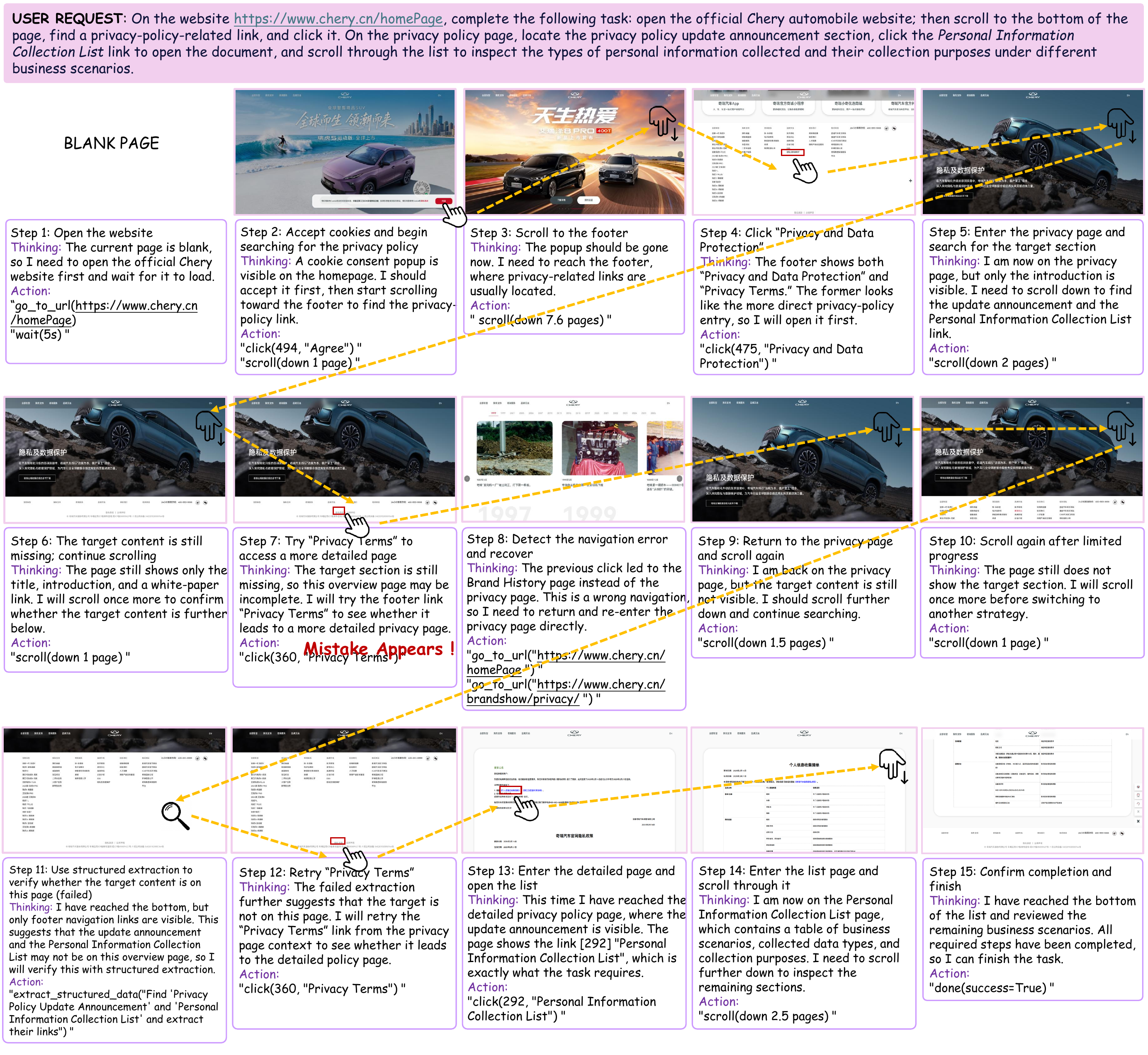}
    \caption{Self-reflective error recovery in long-horizon web navigation. A 15-step trajectory of Wuying-Browser-Agent-9B on a Chery privacy-policy retrieval task. The agent detects an erroneous redirect from the observed browser state, performs corrective recovery, and escalates from scrolling to structured probing before reaching the personal information collection list. For ease of visualization, we display screenshots for all steps to highlight interface transitions; in actual deployment, the agent primarily consumes structured browser state (e.g., DOM-derived representations), while screenshots are used only as optional inputs when visual grounding is needed.}
    \label{fig:reflection_vis}
\end{figure}

\subsubsection{Qualitative Analysis}

Figure~\ref{fig:reflection_vis} presents a representative long-horizon trajectory that illustrates the self-corrective browser behavior of Wuying-Browser-Agent-9B.
The task requires the agent to enter the Chery automotive website, navigate to the privacy-policy page, locate the privacy-policy update announcement, open the personal information collection list, and inspect the categories and purposes of collected personal information.

The trajectory exhibits a clear perceive--act--verify--recover loop rather than a monotonic execution pattern.
In the early stage (Steps 1--5), the agent correctly loads the website, handles the cookie banner, scrolls to the footer, and enters the privacy-related page.
At Step 7, however, it clicks a misleading footer link and is silently redirected to an unrelated brand history page. Instead of assuming progress from the issued action, the agent verifies the resulting browser state against the intended goal and, at Step 8, detects the mismatch from the URL and page content.
It then recovers by re-entering the privacy path.
When repeated scrolling still fails to reveal the target section (Steps 9--10), the agent escalates from low-cost navigation to structured page probing before retrying the correct route.
This eventually allows it to reach the detailed policy page, open the personal information collection list, and complete the final inspection steps (Steps 12--15).

This recovery behavior is consistent with the complementary roles of RUIC-SFT and DAO-GRPO.
RUIC-SFT provides recovery-oriented browser priors through UI-specialized supervision and reflection-rich trajectories, teaching the agent to judge progress from observed environment outcomes rather than from action issuance alone.
DAO-GRPO further strengthens this ability under long-horizon online interaction.
In this example, the successful and unsuccessful paths share a long common prefix, while only a few branch decisions determine whether the agent reaches the target document or remains trapped in an irrelevant page.
This aligns with our divergence-aware online RL design, which concentrates optimization on decisive branch steps such as erroneous redirects, recovery actions, and strategy switches from scrolling to structured extraction.
In addition, the response-level objective optimizes each decision under its actual reconstructed browser context, helping the model rely on the current page state, URL, and visual evidence when deciding whether to continue, recover, or escalate.

Overall, the case study shows that the gain of our method is not merely higher task success, but stronger closed-loop interaction quality.
The model can detect navigation failure, revise its hypothesis, avoid ineffective repetition, and recover toward the original goal under dynamically changing web contexts.




\section{Conclusion}

In this work, we identified the mismatch between success-dominated short-horizon training and real long-horizon browser deployment as the central bottleneck limiting the real-world robustness of web agents. To address this mismatch, we proposed a unified pipeline in which supervision, online optimization, and evaluation are co-designed for the same long-horizon deployment setting. At the supervision level, RUIC-SFT provides a capability-structured initialization by integrating reflection-rich recovery data and specialized UI interaction data through a progressive curriculum, establishing both execution stability and self-correction priors. At the optimization level, DAO-GRPO refines long-horizon decision making under live browser interaction through potential-based reward shaping, divergence-aware step credit assignment, and response-level optimization under dynamically reconstructed contexts. At the evaluation level, BrowserBench serves as a co-designed diagnostic instrument with structured success criteria and capability-aligned semantic tags, enabling fine-grained validation beyond aggregate success rates. Extensive experiments demonstrate that the resulting Wuying-Browser-Agent series establishes a new open-source state of the art across multiple live-web benchmarks while remaining competitive with proprietary agents.

\bibliographystyle{plain}
\bibliography{main}

\clearpage
\clearpage
\thispagestyle{plain}

\begingroup
\setlength{\parindent}{0pt}
\setlength{\parskip}{0.4em}

\phantomsection
\label{sec:contrib_ack}
{\Large\sffamily\bfseries\thupurple{Contributions and Acknowledgments}\par}
\vspace{0.5em}

{\normalsize
All contributors are listed in alphabetical order by their last names.
\par}

\vspace{1em}

\begin{minipage}[t]{0.47\textwidth}
{\bfseries\thupurple{Core Contributors}\par}
\vspace{0.3em}
\begin{itemize}
    \item AIMAE Team
    \item Tianxiang Chen
    \item Yan Cheng
    \item Zhangye Han
    \item Xiaowei Li
    \item Chang Liu
    \item Cheng Liu
    \item Zhongqiang Ma
    \item Long Peng
    \item Xiaobing Tu$^{\dagger}$
    \item Yinggui Wang
    \item Hongliang Wei
    \item Chen Wu
    \item Daiping Xin
    \item Kunyu Zhou
    \item Pengyang Zhou
\end{itemize}

\vspace{1em}
{\bfseries\thupurple{Supervisors}\par}
\vspace{0.3em}
\begin{itemize}
    \item Xiaobing Tu$^{\dagger}$
    \item Yinggui Wang
\end{itemize}
\end{minipage}
\hfill
\begin{minipage}[t]{0.47\textwidth}
{\bfseries\thupurple{Contributors}\par}
\vspace{0.3em}
\begin{itemize}
    \item Peiyuan Chen
    \item Ziyuan Chen
    \item Yutao Deng
    \item Chunyu Dong
    \item Xiangyu Fu
    \item Yicheng Feng
    \item Ruian He
    \item Haochen Li
    \item Miancan Liu
    \item Zhengqin Liu
    \item Wei Peng
    \item \textbf{Jinkui Ren}
    \item Haoyu Tan
    \item Dong Xiao
    \item Rongkun Xue
    \item Shujian Yang
    \item Xianhang Ye
    \item Ziqi Yuan
    \item Ziyang Yu
    \item Linghan Zhang
    \item \textbf{Xiantao Zhang}
    \item Xuanpu Zhao
    \item Yinan Zhao
    \item Zhenghui Zhao
    \item Bin Zhu
    \item Likai Zou
\end{itemize}
\end{minipage}

\vfill
\noindent\rule{0.35\textwidth}{0.4pt}\par
\vspace{0.2em}

{\footnotesize
$^{\dagger}$ Corresponding Author: Xiaobing Tu (\texttt{xiaobing.tuxiaobin@alibaba-inc.com}).
\par}

\endgroup

\clearpage

\end{document}